\documentclass[lettersize,journal]{IEEEtran}
\usepackage{amsmath,amsfonts}
\usepackage{algorithm}
\usepackage{array}
\usepackage[caption=false,font=normalsize,labelfont=sf,textfont=sf]{subfig}
\usepackage{textcomp}
\usepackage{stfloats}
\usepackage{url}
\usepackage{verbatim}
\usepackage{graphicx}
\usepackage{cite}
\usepackage{xcolor}

\newcommand{\bh}[1]{\textcolor{black}{#1}}
\usepackage[utf8]{inputenc} 
\usepackage[T1]{fontenc}    
\usepackage{booktabs}       
\usepackage{amsfonts}       
\usepackage{nicefrac}       
\usepackage{microtype}      
\usepackage{xcolor}         
\usepackage{colortbl}
\usepackage{enumitem}
\usepackage{algpseudocode}
\usepackage{times}
\usepackage{epsfig}
\usepackage{amssymb}
\usepackage{caption}
\usepackage{subcaption}
\usepackage{tabularx}
\usepackage{rotating}
\usepackage{etoolbox}
\usepackage[normalem]{ulem}
\usepackage{multirow}
\usepackage{lipsum}
\usepackage{stmaryrd}
\usepackage{stackengine}
\usepackage{makecell}
\usepackage{pifont}
\usepackage{cancel}
\usepackage{adjustbox}
\usepackage{wrapfig}
\usepackage{bm}

\usepackage{booktabs, multirow} 
\definecolor{roadcolor}{RGB}{234,51,246}
\definecolor{sidewalkcolor}{RGB}{68,8,72}
\definecolor{parkingcolor}{RGB}{241,156,249}
\definecolor{othergroundcolor}{RGB}{160,32,76}
\definecolor{buildingcolor}{RGB}{246,202,69}
\definecolor{carcolor}{RGB}{111,149,238}
\definecolor{truckcolor}{RGB}{74,32,172}
\definecolor{bicyclecolor}{RGB}{136,227,242}
\definecolor{motorcyclecolor}{RGB}{37,59,146}
\definecolor{othervehiclecolor}{RGB}{96,81,242}
\definecolor{vegetationcolor}{RGB}{79, 173, 50}
\definecolor{trunkcolor}{RGB}{126, 65, 22}
\definecolor{terraincolor}{RGB}{171, 238, 105}
\definecolor{personcolor}{RGB}{234, 60, 49}
\definecolor{bicyclistcolor}{RGB}{234, 66, 195}
\definecolor{motorcyclistcolor}{RGB}{138, 42, 90}
\definecolor{fencecolor}{RGB}{238, 128, 69}
\definecolor{polecolor}{RGB}{252, 241, 161}
\definecolor{trafficsigncolor}{RGB}{233, 51, 35}
\definecolor{color1}{RGB}{176, 36, 24}
\definecolor{color2}{RGB}{119,185,0}
\definecolor{color3}{RGB}{0, 0, 200}
\definecolor{colorofteaser}{RGB}{176, 36, 24}

\definecolor{LightGrey}{rgb}{.9,.9,.9}
\definecolor{White}{rgb}{1.,0.,1.}
\definecolor{first}{rgb}{.8,.0,.0}
\definecolor{second}{rgb}{.0,.6,.0}
\definecolor{third}{rgb}{.0,.0,.8}
\definecolor{ceiling}{RGB}{214,  38, 40}   %
\definecolor{floor}{RGB}{43, 160, 4}     %
\definecolor{wall}{RGB}{158, 216, 229}  %
\definecolor{window}{RGB}{114, 158, 206}  %
\definecolor{chair}{RGB}{204, 204, 91}   %
\definecolor{bed}{RGB}{255, 186, 119}  %
\definecolor{sofa}{RGB}{147, 102, 188}  %
\definecolor{table}{RGB}{30, 119, 181}   %
\definecolor{tvs}{RGB}{160, 188, 33}   %
\definecolor{furniture}{RGB}{255, 127, 12}  %
\definecolor{objects}{RGB}{196, 175, 214} %
\definecolor{car}{rgb}{0.39215686, 0.58823529, 0.96078431}
\definecolor{bicycle}{rgb}{0.39215686, 0.90196078, 0.96078431}
\definecolor{motorcycle}{rgb}{0.11764706, 0.23529412, 0.58823529}
\definecolor{truck}{rgb}{0.31372549, 0.11764706, 0.70588235}
\definecolor{other-vehicle}{rgb}{0.39215686, 0.31372549, 0.98039216}
\definecolor{person}{rgb}{1.        , 0.11764706, 0.11764706}
\definecolor{bicyclist}{rgb}{1.        , 0.15686275, 0.78431373}
\definecolor{motorcyclist}{rgb}{0.58823529, 0.11764706, 0.35294118}
\definecolor{road}{rgb}{1.        , 0.        , 1.        }
\definecolor{parking}{rgb}{1.        , 0.58823529, 1.        }
\definecolor{sidewalk}{rgb}{0.29411765, 0.        , 0.29411765}
\definecolor{other-ground}{rgb}{0.68627451, 0.        , 0.29411765}
\definecolor{building}{rgb}{1.        , 0.78431373, 0.        }
\definecolor{fence}{rgb}{1.        , 0.47058824, 0.19607843}
\definecolor{vegetation}{rgb}{0.        , 0.68627451, 0.        }
\definecolor{trunk}{rgb}{0.52941176, 0.23529412, 0.        }
\definecolor{terrain}{rgb}{0.58823529, 0.94117647, 0.31372549}
\definecolor{pole}{rgb}{1.        , 0.94117647, 0.58823529}
\definecolor{traffic-sign}{rgb}{1.        , 0.        , 0.    } 

\definecolor{barrier1}{RGB}{112,128,144}
\definecolor{bicycle1}{RGB}{220,20,60}
\definecolor{bus1}{RGB}{255, 127, 80}
\definecolor{car1}{RGB}{255, 158, 0}
\definecolor{const. veh.1}{RGB}{233, 150, 70}
\definecolor{motorcycle1}{RGB}{255,61,99}
\definecolor{pedestrian1}{RGB}{0,0,230}
\definecolor{traffic cone1}{RGB}{47,79,79}
\definecolor{trailer1}{RGB}{255,140,0}
\definecolor{truck1}{RGB}{255,99,71}
\definecolor{drive. suf.1}{RGB}{0,207,191}
\definecolor{other flat1}{RGB}{175,0,75}
\definecolor{sidewalk1}{RGB}{75,0,75}
\definecolor{terrain1}{RGB}{112,180,60}
\definecolor{manmade1}{RGB}{222,184,135}
\definecolor{vegetation1}{RGB}{0,175,0}

\definecolor{barrier}{RGB}{112,128,144}
\definecolor{bicycle}{RGB}{220,20,60}
\definecolor{bus}{RGB}{255, 127, 80}
\definecolor{car}{RGB}{255, 158, 0}
\definecolor{const. veh.}{RGB}{233, 150, 70}
\definecolor{motorcycle}{RGB}{255,61,99}
\definecolor{pedestrian}{RGB}{0,0,230}
\definecolor{traffic cone}{RGB}{47,79,79}
\definecolor{trailer}{RGB}{255,140,0}
\definecolor{truck}{RGB}{255,99,71}
\definecolor{drive. suf.}{RGB}{0,207,191}
\definecolor{other flat}{RGB}{175,0,75}
\definecolor{sidewalk}{RGB}{75,0,75}
\definecolor{terrain}{RGB}{112,180,60}
\definecolor{manmade}{RGB}{222,184,135}
\definecolor{vegetation}{RGB}{0,175,0}
\definecolor{y}{HTML}{00994C}
\definecolor{b}{rgb}{0.31372549, 0.11764706, 0.70588235}

\usepackage{soul}

\newcommand{\re}[1]{\textcolor{black}{#1}}
\newcommand{\rcap}{\captionsetup{labelfont={color=black}, textfont={color=black}}}

\begin{document}

\title{
\textcolor{other flat1}{Occ}\textcolor{b}{Scene}: Semantic \textcolor{other flat1}{Occ}upancy-based \\Cross-task Mutual Learning for 3D \textcolor{b}{Scene} Generation 
}

\author{Bohan Li,~Xin Jin,~\IEEEmembership{Member, IEEE},~Jianan Wang,~Yukai Shi,~Yasheng Sun,~Xiaofeng Wang,~Zhuang Ma,\\~Baao Xie,~\IEEEmembership{Member, IEEE},~Chao Ma,~\IEEEmembership{Member, IEEE},~Xiaokang Yang,~\IEEEmembership{Fellow, IEEE},~Wenjun Zeng,~\IEEEmembership{Fellow, IEEE}
\vspace{-7pt}
    \thanks{Bohan Li is with the School of Electronic Information and Electrical Engineering, Shanghai Jiao Tong University, Shanghai, China, and Ningbo Institute of Digital Twin, Eastern Institute of Technology, Ningbo, China (e-mail: bohan\_li@sjtu.edu.cn). Xiaokang Yang is a distinguished professor, and Chao Ma is an associate professor at the School of Electronic Information and Electrical Engineering, Shanghai Jiao Tong University, Shanghai, China.}	
    \thanks{Jianan Wang, Yukai Shi, Yasheng Sun, and Xiaofeng Wang are with the Astribot, Shenzhen, China.}
    \thanks{Zhuang Ma is with PhiGent Robotics, Beijing, China.}
    \thanks{ Wenjun Zeng is a chair professor, Xin Jin (corresponding author) is an assistant professor, and Baao Xie is a postdoctoral researcher at the Ningbo Institute of Digital Twin, Eastern Institute of Technology, Ningbo, China (e-mail: jinxin@eitech.edu.cn).}
}

\markboth{SUBMITTED TO IEEE Transactions on Pattern Analysis and Machine Intelligence}%
{Shell \MakeLowercase{\textit{et al.}}: A Sample Article Using IEEEtran.cls for IEEE Journals}

\maketitle

\begin{abstract}
Recent diffusion models have demonstrated remarkable performance in both 3D scene generation and perception tasks. Nevertheless, existing methods typically separate these two processes, acting as a data augmenter to generate synthetic data for downstream perception tasks. In this work, we propose \textbf{OccScene}, a novel mutual learning paradigm that integrates fine-grained 3D perception and high-quality generation in a unified framework, achieving a cross-task win-win effect. OccScene generates new and consistent 3D realistic scenes only depending on text prompts, guided with semantic occupancy in a joint-training diffusion framework. To align the occupancy with the diffusion latent, a Mamba-based Dual Alignment module is introduced to incorporate fine-grained semantics and geometry as perception priors. Within OccScene, the perception module can be effectively improved with customized and diverse generated scenes, while the perception priors in return enhance the generation performance for mutual benefits. Extensive experiments show that OccScene achieves realistic 3D scene generation in broad indoor and outdoor scenarios, while concurrently boosting the perception models to achieve substantial performance improvements in the 3D perception task of semantic occupancy prediction.
\end{abstract}


\begin{figure*}[!t]
\vspace{-0pt}
\centering
\includegraphics[width=1.0\linewidth]{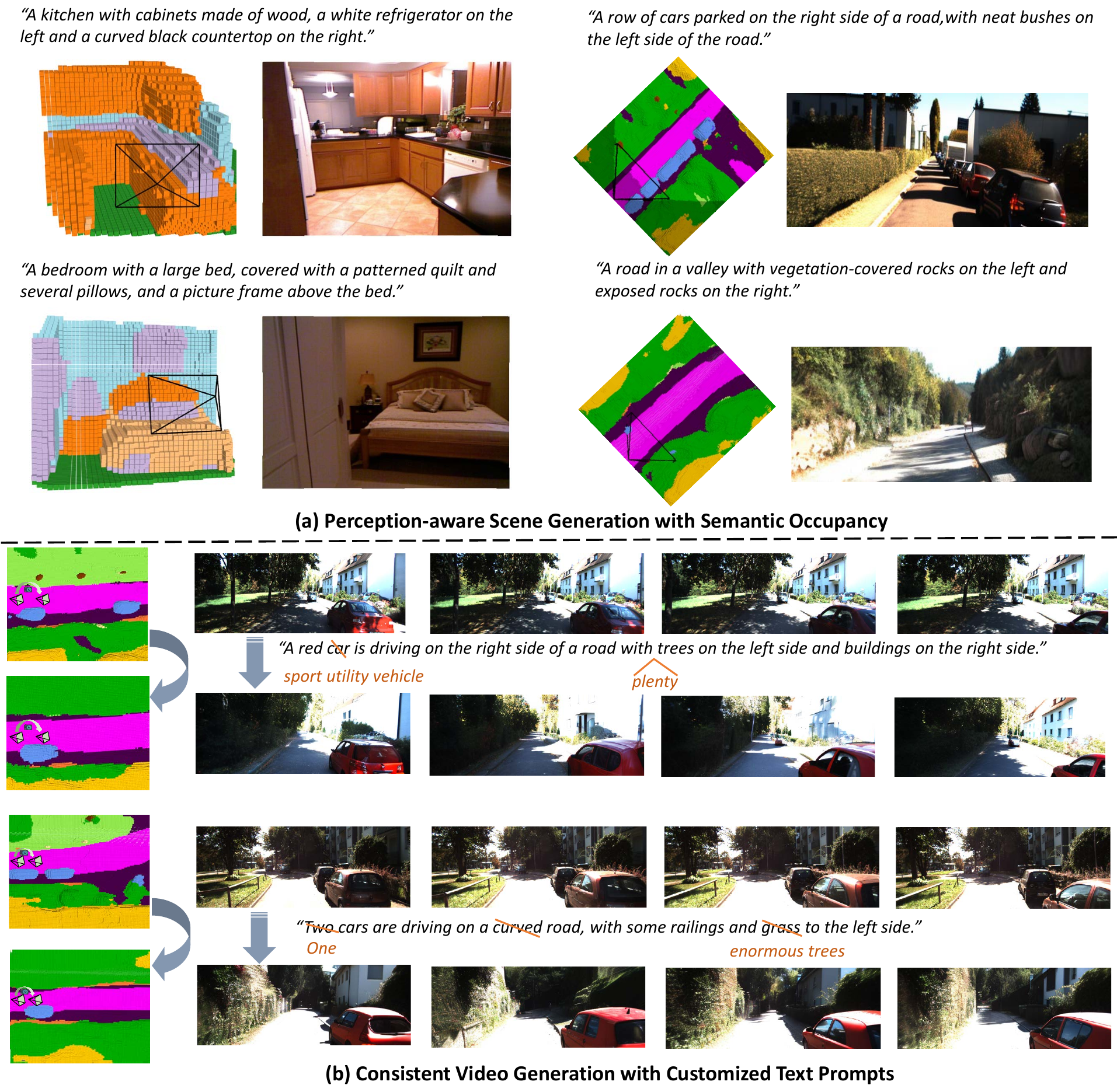}
\vspace{-0pt}
\caption{OccScene synthesizes realistic scene generation in RGB and semantic occupancy pairs from customized text prompts. (a): example of indoor room-scale and outdoor autonomous driving scene generation with perception {awareness}; (b): occupancy-based consistent video generation and editing with controllable customized text prompts.}
\label{teaser}
\vspace{-0pt}
\end{figure*}

\section{Introduction}\label{sec_intro}
\IEEEPARstart{T}{he} effectiveness of 3D perception models significantly relies on large-scale data collection with precisely annotated labels~\cite{rist2021semantic,liao2022kitti,li2021anisotropic,li2023front,li2023stereoscene}. However, obtaining these datasets requires substantial resources and manual effort.
Recent advancements in generative diffusion models~\cite{luo2021diffusion,rombach2022high,jiang2022conditional,li2024time} have made it possible to generate high-fidelity images, thereby enabling the training with synthetic datasets for out-of-distribution (OOD) perception generalization~\cite{yang2023bevcontrol,swerdlow2024street,wang2023drivedreamer}.
These datasets, generated by cutting-edge models, have proven effective in enhancing the performance of 2D object-level downstream tasks such as object detection~\cite{bowles2018gan,chen2023integrating,wang2024detdiffusion}, classification~\cite{he2022synthetic,moller2023prompt}, and segmentation~\cite{li2023open,wu2023datasetdm}.
Despite the remarkable achievements of existing generative frameworks for 2D object-level tasks, generating scene-level 3D data with realistic layout and geometry still remains challenging due to the complexity and diversity of real-world scenes modeling~\cite{yang2023bevcontrol,swerdlow2024street,wang2023drivedreamer,zhao2024drivedreamer,gao2023magicdrive,wang2023driving}.

Recently, some works attempt to incorporate prior knowledge from 3D ground-truth (GT) labels (e.g., 3D bounding boxes and BEV maps) to assist the generation of realistic scenes in the inference process, thereby improving downstream tasks with synthetic data~\cite{wang2023drivedreamer,gao2023magicdrive,wang2023driving}. 
Specifically, 
DriveDreamer~\cite{wang2023drivedreamer} and DriveDreamer-2~\cite{zhao2024drivedreamer} incorporate ground-truth road structure information for driving video generation to improve downstream perception tasks.
MagicDrive~\cite{gao2023magicdrive} proposes to leverage 3D geometry information from GT labels (e.g., camera poses, road maps, and 3D bounding boxes) to synthesize new data for perception task enhancement.
Besides, these GT-based methods typically consider the generation and perception processes separately, and trivially leverage the pre-trained generator as a data augmenter for improving perception tasks. These issues inevitably pose several challenges:
(I) \textbf{Limited Flexibility.} Generating 3D scenes based on ground-truth labels like~\cite{wang2023driving,gao2023magicdrive} in the inference process depends on high annotation costs and hardly generates diverse corner cases; (II) \textbf{Insufficient Constraints.} Complex real-world scene generation requires pixel-level fine-grained semantics and geometry guidance, but the existing region-level coarse prior (e.g., 3D bounding boxes) used in~\cite{wang2023drivedreamer,gao2023magicdrive,wang2023driving} struggle to provide sufficient context; (III) \textbf{Unclear Goals.} The previous scene generation works are typically subjective quality-driven and perception-irrelevant, which makes the generated data less valuable for downstream complex perception tasks like in autonomous driving and robot navigation.\looseness=-1

To address these challenges, we propose \textbf{OccScene}, a novel mutual learning paradigm in 3D scene generation that unifies the two tasks of semantic occupancy~\cite{cao2022monoscene,li2023stereoscene} prediction/perception and text-driven controllable generation.
\bh{Instead of enhancing performance for a single generation task with independent
learning, OccScene enables cross-task collaboration for mutual benefits throughout a joint learning scheme.}
In this way, unlike previous methods that rely on ground-truth labels~\cite{higgins2016beta,preechakul2022diffusion,jin2024closed}, OccScene generates realistic images or videos and their corresponding semantic occupancy simultaneously within a unified framework via only text prompts.\looseness=-1

\bh{The effectiveness of the proposed learning strategy may not be immediately clear. One might ask: Where does the additional knowledge come from? Why does the strategy converge to an optimal solution instead of resulting in failure, akin to 'the blind leading the blind'? 
As pointed out in~\cite{zhang2018deep,yang2021mutualnet,wang2023spatial}, some intuition about these questions can be gleaned with the following factors: 
Both the learners involved in the perception and generation tasks are primarily guided by conventional supervised learning losses, leading to a general improvement in performance. Through supervised learning, both learners quickly produce the correct labels for each training instance. However, because they are focused on distinct tasks-perception and generation, they develop different representations of the data. As a result, their comprehension and predictions for the same 3D scenario differ. These differing representations contribute the additional information necessary for cross-task mutual learning.
In mutual learning, the perception and generation learners refine their collective understanding of the 3D scenario. By comparing and aligning their representations of the scenario for each training instance, each learner increases its posterior entropy. This increase in posterior entropy enables both learners to converge towards a more robust solution, characterized by a flatter minima, which leads to better generalization on testing data.}\looseness=-1

As shown in Figure~\ref{teaser}, OccScene could generate high-quality RGB-Occupancy pairs for indoor and outdoor scenes. Furthermore, OccScene enables occupancy-based cross-view video generation and consistent editing.
Technically, we introduce a joint learning scheme to improve the perception and generation performance concurrently during the diffusion process.
This framework enhances the perception model by utilizing customized generation results and noisy input images with varying information capacities during the generative process. 
To efficiently provide occupancy-based priors for the diffusion model, we propose a Mamba-based Dual Alignment (MDA) module with the linear-complexity operator.
This module effectively aligns the semantic occupancy and the diffusion latent with the camera parameters, thereby ensuring cross-view generation consistency with camera trajectory awareness and providing fine-grained semantics and geometry guidance with aligned contextual information. The main contributions of this paper are summarized as follows:

\begin{itemize}

\item We present a novel generation paradigm that harmonizes 3D scene perception and generation, enabling mutual benefits in a joint diffusion process.

\item To enhance generation performance with the perception model, we introduce a Mamba-based Dual Alignment module to facilitate cross-view consistency through camera trajectory awareness, and incorporate fine-grained geometry and semantics with aligned context.

\item {To improve perception performance within the generative framework, we incorporate the perception model into the generation process for joint learning and customized data augmentation with text-driven diverse scene generation.} 
\vspace{-0pt}
\end{itemize}

Extensive experiments demonstrate that OccScene achieves high-fidelity scene data synthesis and effectively improves the perception model as a plug-and-play training strategy. For the 3D perception task of semantic occupancy prediction, 
our method shows that the use of generated synthetic data leads to significant improvements in performance.

\section{Related Work}

\subsection{Diffusion Models for Scene Generation}

Diffusion Models, a recently established class of generative models grounded in non-equilibrium thermodynamics theory~\cite{sohl2015deep}, which delineate empirical data distributions through an iterative noise reduction mechanism~\cite{shao2022diffustereo}, closely paralleling score-based generative models that rely on Langevin dynamics leveraging inferred data distribution gradients~\cite{jiang2022conditional}.
Diffusion models have significantly advanced fields such as text-to-image generation~\cite{rombach2022high, huang2023t2i, zhang2023adding, tumanyan2023plug} and controllable video generation~\cite{wu2023tune,videoworldsimulators2024,du2024learning,liu2024fetv}. These models have also been developed to support downstream applications, notably in autonomous driving scene generation~\cite{wang2023drivedreamer,zhao2024drivedreamer,wang2023driving,huang2024subjectdrive,swerdlow2024street}. 
One class of the driving generators is based on NeRF and Gaussian Splatting~\cite{wu2023mars,yan2024street,yariv2024diverse,xiangli2022bungeenerf}, which suffer from poor diversity. Another class involves world models or world generators, with notable examples including DriveDreamer~\cite{wang2023drivedreamer}, BEVGen~\cite{swerdlow2024streetview}, Panacea~\cite{wen2023panacea}, Drive-WM~\cite{wang2023driving}, etc.
\re{Recent advancements in LiDAR generation and semantic scene generation also highlighted the potential of diffusion models. In LiDiff~\cite{nunes2024cvpr}, a diffusion model is adapted to directly process sparse 3D LiDAR point clouds for scene completion, achieving superior detail recovery compared to range-image-based methods. SemCity~\cite{lee2024semcity} introduces a triplane diffusion model for semantic scene generation to address data sparsity challenges in real-world outdoor environments.}
Recently, some studies have utilized generated data to enhance downstream perception models.
DetDiffusion~\cite{wang2024detdiffusion} introduces perception-aware loss and attributes to improve the quality of the generation images for 2D object detection.
To facilitate realistic scene generation, MagicDrive~\cite{gao2023magicdrive} leverages 3D geometry information from ground-truth labels.
However, this method depends heavily on ground-truth labels in the inference process and faces significant challenges in generating flexible and generalizable real-world scenes.\looseness=-1

\subsection{Semantic Occupancy Prediction}
\re{Semantic occupancy prediction(SOP) is a dense 3D perception task that unifies semantic segmentation with scene completion~\cite{roldao20223d}. Prior research has extensively employed LiDAR to capitalize on its 3D geometric data capabilities~\cite{song2017semantic,roldao2020lmscnet,yan2021sparse,xia2023scpnet,cao2024pasco}.
SSCNet~\cite{song2017semantic} pioneers an end-to-end 3D convolutional network for joint occupancy and semantic prediction from a single depth image. LMSCNet~\cite{roldao2020lmscnet} introduces lightweight multiscale architectures for efficient scene completion. JS3CNet~\cite{yan2021sparse} leverages contextual shape priors from sequential LiDAR data to enhance sparse point cloud segmentation. SCPNet~\cite{xia2023scpnet} improves robustness through innovative sub-network designs and knowledge distillation. PaSCo~\cite{cao2024pasco} extends semantic scene completion to panoptic scene completion, introducing uncertainty awareness critical for robotics applications.
Recent self-supervised methods have also advanced occupancy prediction~\cite{cao2023scenerf,wimbauer2023behind,huang2024selfocc}. SceneRF~\cite{cao2023scenerf} employs neural radiance fields (NeRF) with explicit depth optimization for monocular 3D reconstruction, excelling in novel depth synthesis. Behind the Scenes~\cite{wimbauer2023behind} proposes a density field-based approach for volumetric occupancy prediction, effectively handling occlusions. SelfOcc~\cite{huang2024selfocc} introduces a self-supervised framework using video sequences, eliminating the need for voxel annotations by leveraging signed distance fields (SDF) and multi-view constraints.}
Moreover, camera-driven 3D SOP has garnered significant interest due to the affordability and mobility of camera systems~\cite{silberman2012indoor,yan2021sparse,cheng2021s3cnet,wu2020scfusion,cao2022monoscene,li2023voxformer,li2024hierarchical}. MonoScene~\cite{cao2022monoscene} propose to infer both geometry and semantics from a single RGB image via 2D-to-3D feature projection, which sparked a wave of advancements in camera-based scene understanding~\cite{huang2023tri,zhang2023occformer,wei2023surroundocc,li2023stereoscene}.
TPVFormer~\cite{huang2023tri} innovates with a tri-perspective framework for detailed 3D scene depiction. 
OccFormer~\cite{zhang2023occformer} devises a dual-path transformer to handle dense 3D feature processing for semantic occupancy. 
SurroundOcc~\cite{wei2023surroundocc} introduces multi-view image inputs for enhanced occupancy estimation. 
Pioneered by VPD~\cite{li2024time}, conditional diffusion models are leveraged for 3D perception tasks including multi-view stereo and semantic occupancy prediction.
However, how to use the powerful generative models to produce high-quality data pairs and thus improve perception remains unexplored.

The methods discussed above typically separate the perception and generation processes, resulting in limited flexibility and unclear goals -- the generation relies on ground-truth annotations and the generated data may be useless for perception. In this work, we propose incorporating perception models into the generation framework to establish a joint optimization for mutual benefits.

\begin{figure*}[!h]
\rcap
\vspace{-0pt}
\centering
\includegraphics[width=1.0\linewidth]{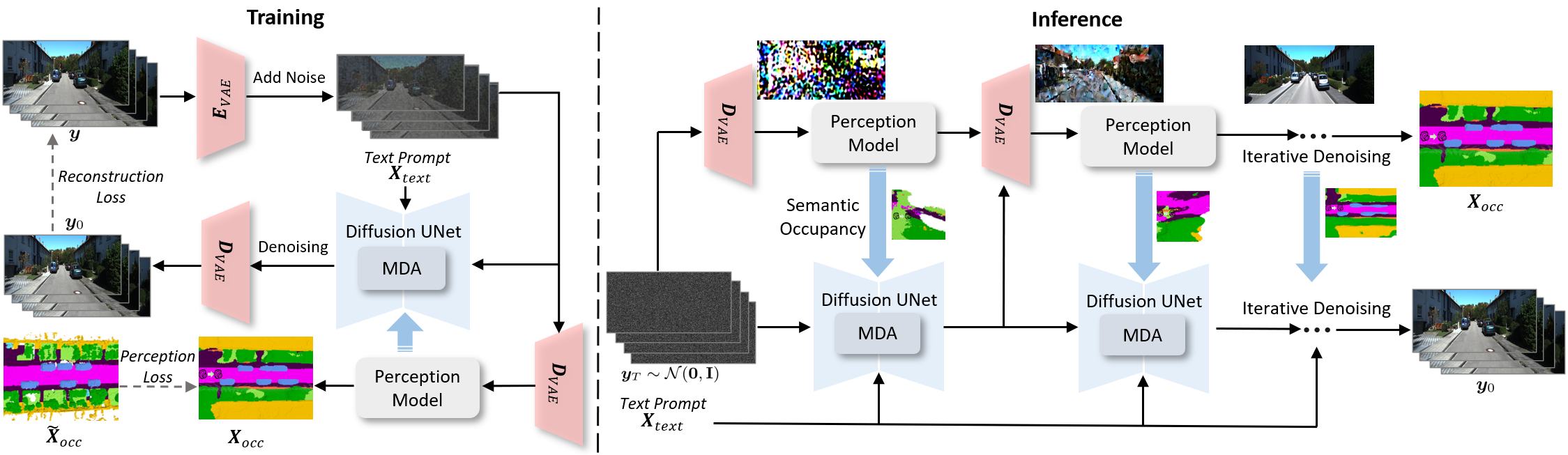}
\vspace{-0pt}
\caption{
\re{Overview of the proposed OccScene. The framework involves the concurrent training of the perception model and the generative diffusion UNet. A reconstruction loss is leveraged between the ground-truth images or videos $\bm{y}$ and generation results $\bm{y}_0$, while a perception loss is adopted between the ground-truth semantic occupancy $\bm{\tilde{X}}_{occ}$ and predicted semantic occupancy $\bm{X}_{occ}$.
During the inference process, OccScene takes Gaussian noise $\bm{y}_T \sim \mathcal{N}(\mathbf{0}, \mathbf{I})$ and conditional text prompt $\bm{X}_{text}$ as inputs, facilitating the simultaneous generation of images or videos $\bm{y}_0$ and their associated semantic occupancy $\bm{X}_{occ}$. 
Within OccScene, we introduce a Mamba-based Dual Alignment (MDA) module to sequentially align the semantic occupancy and the diffusion latent with camera trajectory awareness.}
}
\label{overall}
\vspace{-0pt}
\end{figure*}

\vspace{-0pt}
\begin{algorithm*}[!t]
  \caption{Training the generation model $f_\theta$ and the perception model $f_\delta$ simultaneously.} \label{alg:training}
  \small
  \begin{algorithmic}[1]
    \Repeat
      \State $(\bm{X}_{(occ,text)}, \bm{y}_0) \sim p(\bm{X}_{(occ,text)}, \bm{y})$
      \State $\bar{\alpha}_t \sim p(\bar{\alpha}_t)$
      \State $\bm{\epsilon}\sim\mathcal{N}(\mathbf{0},\mathbf{I})$
      
      \State $\bm{X}_{occ} = f_\delta(\bm{y})$
      \State Take a gradient descent step on
      \State $ \quad {\nabla_\theta\left\|f_\theta\left(  \bm{X}_{(occ,text)} , \sqrt{\bar{\alpha}_t} \bm{y}_0+\sqrt{1-\bar{\alpha}_t} \epsilon, \bar{\alpha}_t  \right)-\epsilon\right\|} + {   \nabla_\delta \left\|  \bm{\tilde{X}}_{occ} - \bm{X}_{occ}  \right\|}$ 
    \Until{converged}
  \end{algorithmic}
\end{algorithm*}

\begin{algorithm*}[!t]
  \caption{Consistency constrained inference in $T$ iterative steps.} \label{alg:sampling}
  \small
  \begin{algorithmic}[1]
    \State $  \bm{y}_T \sim \mathcal{N}(\mathbf{0}, \mathbf{I})  $
    
    \For{$t=T, \dotsc, 1$}
      \State $\mathbf{z} \sim \mathcal{N}(\mathbf{0}, \mathbf{I})$ if $t > 1$, else $\mathbf{z} = \mathbf{0}$
      \State $\bm{X}_{occ} = f_\delta (\bm{y}_t)$
      \State $\bm{y}_{t-1} = \frac{1}{\sqrt{\alpha_t}}\left( \bm{y}_t - \frac{1-\alpha_t}{\sqrt{1-\bar{\alpha}_t}} f_\theta( \bm{X}_{occ,text}, \bm{y}_t, \bar{\alpha}_t  ) \right) + \sqrt{1 - \alpha_t} \mathbf{z}$
    \EndFor
    \State \textbf{return} $\bm{y}_0$, $\bm{X}_{occ}$
  \end{algorithmic}
\end{algorithm*}

\section{Methodology}\label{sec_met}
The overview of OccScene is illustrated in Figure~\ref{overall}, which simultaneously generates realistic scene images or videos and their corresponding semantic occupancy.
Instead of utilizing prior knowledge from ground-truth labels in the inference process~\cite{gao2023magicdrive,wang2023driving,zhao2024drivedreamer}, OccScene generates multi-modal results (RGB \& Occupancy) synchronously within a unified framework only via customized text prompts.
In detail, we first introduce the preliminaries in Section~\ref{sec_pre} and present the joint denoising diffusion scheme in Section~\ref{sec_sym}.
The Mamba-based Dual Alignment (MDA) module is illustrated in Section~\ref{sec_cam}. The analysis on the benefits of cross-task mutual learning is presented in Section~\ref{sec_ana}.

\subsection{Preliminaries}\label{sec_pre}

The standard generative diffusion models aim to establish one-to-many mappings with a forward and reverse process~\cite{saharia2022image}.
In the forward process, the input image $\bm{y}_{0}$ is progressively corrupted to $\bm{y}_{T} \sim \mathcal{N}(\mathbf{0}, \mathbf{I})$ in ${T}$ time steps following a discrete-time Markov chain. 
The distributions of intermediate steps can be characterized by marginalizing:
$
q\left(\bm{y}_t \mid \bm{y}_0\right)=\mathcal{N}\left(\bm{y}_t\mid\sqrt{\bar{\alpha}_t} \bm{y}_0,\left(1-\bar{\alpha}_t\right) \mathbf{I}\right)
$,
where $\bar{\alpha}_t=\prod_{i=1}^t \alpha_i$ and $\alpha_t$ is a pre-defined coefficient. $\mathcal{N}$ and $\mathbf{I}$ denote the normal distribution and the identity matrix, respectively.
In the reverse process, a diffusion neural network such as UNet estimates a corresponding image to approximate the input image $\bm{y}_{0}$ from noisy input $\bm{y}_{T}$ and conditions $\bm{X}$ which are optionally provided to guide the estimation process. 
 Each step of the reverse process can be defined as conditional distribution transition~\cite{saharia2022image}, which is formulated as:
$
 p_\theta\left(\bm{y}_{0: T} \mid \bm{X}\right)=p\left(\bm{y}_T\right) \prod_{t=1}^{T} p_\theta\left(\bm{y}_{t-1} \mid \bm{y}_t, \bm{X}\right)
$,
where $p_\theta$ represents the reverse function and $\bm{X}$ denotes the conditions of the diffusion model.

Stable Diffusion (SD)~\cite{rombach2022high}, as a Latent Diffusion Model (LDM), features an efficient pipeline for Text-to-Image (T2I) synthesis. The process encodes the input images into a latent space using a Variational AutoEncoder (VAE) for compression, and learns the diffusion process in the latent space. A pre-trained CLIP encoder is utilized to integrate text prompts as conditions. We use SD in our work as a strong generative backbone and baseline to be compared.

\subsection{\bh{Joint Perception-Generation Diffusion Scheme}}\label{sec_sym}
To facilitate the mutual benefits of perception and generation, we introduce a joint scheme to unify two tasks of semantic occupancy prediction and text-driven generation into a single diffusion process. Throughout the joint learning scheme, OccScene enables cross-task collaboration for general performance improvements.\looseness=-1

\noindent\textbf{Training Process.} 
As illustrated in Figure~\ref{overall}, the generative Diffusion UNet and the perception model are learned together during the training process.
Specifically, the input images are first compressed with the VAE encoder $\bm{E}_{\mathrm{VAE}}$, followed by noise injection to yield a latent feature $\bm{L}$. The latent feature $\bm{L}$ is then separately fed into the diffusion UNet for denoising and the VAE decoder $\bm{D}_{\mathrm{VAE}}$ to produce noisy images. 
To condition the diffusion UNet with fine-grained semantics and geometry, the perception model takes the noisy images as inputs to predict the occupancy grids $\bm{X}_{occ}$, which are leveraged as the additional condition to constrain the diffusion UNet.
In our implementation, the camera-based semantic occupancy prediction network~\cite{cao2022monoscene, huang2023tri} with pre-trained weights is utilized as the perception model.

To facilitate stable and robust learning, we adopt a two-stage training schedule: (I) Freeze the weights of the perception model and train the diffusion UNet with conditional guidance to generate realistic scenes; (II) Train the diffusion UNet and the perception model together for mutual benefits.
In the first stage, the diffusion UNet learns to fit the specific training data, thereby generating diverse realistic images to enhance the perception model for the next training stage.
In the second stage, to mitigate the impact of the noise added to the input images, we supervise the perception model according to the varying scales of $\bar{\alpha}_t$, corresponding to different time-steps according to Denoising Diffusion Probabilistic Models~\cite{ho2020denoising}.
In this way, the training scheme counterbalances the noise component in the input images, thereby ensuring the stability and utility of the supervision.
The overall loss function is mathematically represented as follows, which combines the perception-aware loss $\mathcal{L}_p$ with the foundational reconstruction loss $\mathcal{L}_{L D M}$ of the Latent Diffusion Model (LDM):
\begin{equation}
\mathcal{L}=\mathcal{L}_{L D M}+\sqrt{\bar{\alpha}_t} \mathcal{L}_p
,
\end{equation}
where $\sqrt{\bar{\alpha}_t}$ is leveraged to emphasize the supervision with low noise levels (i.e., small time-step) and reduce the impact with high noise levels (i.e., large time-step).  
We implement the perception loss $\mathcal{L}_{p}$ following the MonoScene~\cite{cao2022monoscene} for semantic occupancy prediction. 
Standard semantic loss $\mathcal{L}_{\text{sem}}$ and geometry loss $\mathcal{L}_{\text{geo}}$ are leveraged for semantic and geometry supervision, while an extra class weighting loss $\mathcal{L}_{ce}$ is also added.
The overall learning objective of this framework is formulated as:
\begin{equation}
   { \mathcal{L}_{p} =  \mathcal{\lambda}_{ce} \mathcal{L}_{ce} + \mathcal{\lambda}_{sem} \mathcal{L}_{\text{sem}}+ \mathcal{\lambda}_{geo} \mathcal{L}_{\text{geo}} , }
\end{equation}
where several $\lambda$s are balancing coefficients.

We present the algorithm details of the training process in Algorithm~\ref{alg:training}.
In the algorithm, $\bm{X}_{(occ,text)}$ denotes the conditions, including semantic occupancy $\bm{X}_{occ}$ and text prompt $\bm{X}_{text}$. 
$f_\theta$ and $f_\delta$ represent the generation model and the perception model, respectively. 
$ \bm{\epsilon}  \sim\mathcal{N}(\mathbf{0},\mathbf{I})$ denotes the Gaussian noise. 
In the training process, the perception model and the generative backbone are jointly learned to achieve a win-win effect.
We leverage a pre-trained perception model $f_\delta$ to encode input noisy images and predict semantic occupancy grids to condition the diffusion UNet $f_\theta$.

\noindent\textbf{Inference Process.} As presented in Figure~\ref{overall}, during the inference process, OccScene generates images or videos along with their corresponding semantic occupancy simultaneously.
The framework takes Gaussian noise $\bm{\epsilon}  \sim\mathcal{N}(\mathbf{0},\mathbf{I})$ as input, and leverages a customized text prompt $\bm{X}_{text}$ as the condition.
 In each inference iteration, the perception model takes noisy images decompressed from the VAE decoder $\bm{D}_{\mathrm{VAE}}$ as input and predicts the semantic occupancy grids $\bm{X}_{occ}$ as the additional conditions for the diffusion UNet, thereby improving generation quality and ensure video consistency.
During the iterative inference process, the predicted images become clearer and more informative, resulting in more complete and accurate semantic occupancy grids. These enhanced occupancy grids provide a more specific semantic and geometric context, thereby constraining and refining the generative inference process.

We further present the algorithm details of the inference process in Algorithm~\ref{alg:sampling}.
In the inference process, the framework produces images or videos and corresponding semantic occupancy synchronously. 
The semantic occupancy $\bm{X}_{occ}$, predicted by the perception model, conditions the diffusion UNet to improve generation quality and ensure video consistency.

The occupancy-based generation facilitates fine-grained cross-view control by extending the single-view prompt editing technology~\cite{hertz2022prompt}.
Given an edited text prompt, we 
modify the cross-attention layers for pixel-to-text interaction after incorporating semantic occupancy. 
Please refer to the Supplementary Material for more details about the editing process and Section~\ref{sec_cam} for semantic occupancy incorporation.

\subsection{\bh{Mamba-based Dual Alignment} }\label{sec_cam}

\begin{figure*}[!ht]
\vspace{-0pt}
\hsize=\textwidth %
\centering
\includegraphics[width=0.99\textwidth]{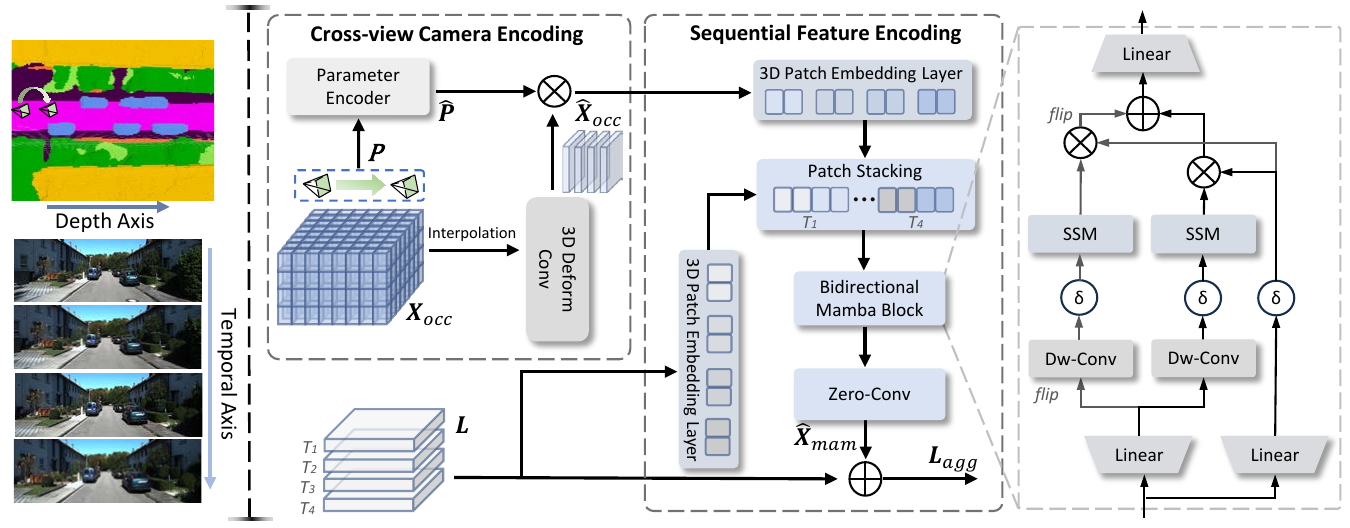}
\vspace{-0pt}
\caption{The architecture of the proposed \bh{Mamba-based Dual Alignment (MDA) module for occupancy-based constraint condition, which is mainly composed of the Cross-view Camera Encoding and the Mamba-based Sequential Feature Encoding}.
The Cross-view Camera Encoding incorporates camera parameters $\bm{P}$ with the semantic occupancy $\bm{X}_{occ}$ for camera trajectory awareness, while the Mamba-based Sequential Feature Encoding processes the semantic occupancy feature $\bm{\widehat{X}}_{occ}$ along the depth dimension and the latent feature $\bm{L}$ along the temporal dimension ($T_1, T_2, \dots $) with the bidirectional mamba block for context alignment.\looseness=-1
}
\label{module}
\vspace{-0pt}
\end{figure*}

To condition the diffusion UNet with occupancy-based constraint, we propose to sequentially align the semantic occupancy $\bm{X}_{occ}$ and the diffusion latent feature $\bm{L}$ with the camera parameter $P$, which is shown in Figure~\ref{module}.
Specifically, to ensure cross-view video consistency, we present \bh{Cross-view Camera Encoding} to incorporate camera parameters with the semantic occupancy for camera trajectory awareness.
Moreover, to align the semantic occupancy $\bm{X}_{occ}$ with the latent features $\bm{L}$, we introduce \bh{Mamba-based Sequential Feature Encoding}, which processes the semantic occupancy along the depth dimension and the latent feature along the temporal dimension with bidirectional mamba block for context alignment.

\noindent\textbf{Cross-view Camera Encoding for View Consistency.}
The occupancy grids generated by the semantic occupancy prediction networks are sufficient to describe a large scene (e.g., $51.2m \times 51.2m \times 6.4m$ in SemanticKITTI). 
To save computational resources, we only utilize the occupancy grid predicted from the first key-frame and slide the camera viewpoint to generate camera trajectory-aware occupancy features. 

Given a latent feature $\bm{L} \in C_L \times N \times H_L \times W_L$ with $N$ video frames and a semantic occupancy grid $\bm{X}_{occ} \in 1\times D\times H_{occ} \times W_{occ}$ predicted from the first key-frame, we aim to encode the occupancy-based features $\bm{\widehat{X}}_{occ}^{i}$ with the camera parameter $\bm{P}^{i}$ for $i^{th}$ view ($i \in (0, N-1)$). 
In this way, the occupancy-based features $\bm{\widehat{X}}_{occ}$ of $N$ frames are incorporated with the corresponding camera trajectory.
Specifically, to encode distinct camera parameters with the key-frame semantic occupancy, 
we feed the camera parameters $\bm{P}^{i}$ (including intrinsic and extrinsic parameters) to a {Parameter Encoder} as:

\vspace{-0pt}
\begin{equation} \label{eq1}
\bm{\widehat{P}}^{i} = \sigma \left ( \operatorname{Conv}  (\operatorname{Reshape} (\operatorname{FC}(\bm{P}^{i}))) \right ) ,
\end{equation}
where $\operatorname{Conv}$ and $\operatorname{FC}$ are convolutions and fully-connected layers, whereas $\sigma$ and $\operatorname{Reshape}$ represent sigmoid function and reshape operation, respectively. 
Next, we interpolate $\bm{X}_{occ}$ to align with the latent feature $\bm{L}$ on the spatial dimension and leverage deformable 3D convolution to generate dynamic occupancy volume for $i^{th}$ camera view, which is multiplied with the encoded camera parameters $\widehat{P}^{i}$ for camera-awareness:

\vspace{-0pt}
\begin{equation} \label{eqca2}
\begin{split}
\bm{\widehat{X}}_{occ}^{i} = \sum_{k=1}^{K_w} w_k \cdot   \bm{X}_{occ}(\mathrm{p}+\mathrm{p}_k+\Delta \mathrm{p}_k) \cdot \bm{\widehat{P}}^{i},
\end{split}
\end{equation}
where $K_w$ represents the number of points in the deformable sampling process, and $w_k$ denotes the spatial feature weight. $\Delta \mathrm{p}_k$ denotes the additional offset in the sampling grid, which adaptively adjusts sampling location $\mathrm{p}+\mathrm{p}_k$.
In this way, $\bm{\widehat{X}}_{occ}^{i}$ encodes specific semantic occupancy features with corresponding camera parameters awareness.

To facilitate video generation of $N$ frames, the camera parameters from different viewpoints are encoded separately, and different deformable 3D convolutions without shared weights are leveraged to generate the corresponding occupancy features.
In this way, the occupancy-based features $\bm{\widehat{X}}_{occ}$ of $N$ frames correspond to the latent features $L$ along the temporal axis.
Note that for single-view image generation, we utilize the same implementation with $N=1$.

\noindent\textbf{Sequential Feature Encoding for Contextual Alignment.}
The semantic occupancy feature $\bm{\widehat{X}}_{occ}$ consists of $N$ semantic maps concatenated in the depth dimension, while the latent feature $\bm{L}$ consists of $N$ video frame features in the temporal dimension. 
To align them for reliable feature encoding, we propose to sequentially scan $\bm{\widehat{X}}_{occ}$ along the depth axis and the latent feature $\bm{L}$ along the temporal dimension with the bidirectional mamba block.

As shown in Figure~\ref{module}, we first project the semantic occupancy feature $\bm{\widehat{X}}_{occ}$ and the latent feature $\bm{L}$ into non-overlapping spatio-temporal patches before feeding into the mamba block. 
The mamba block draws inspiration from the State Space Models (SSMs)~\cite{gu2023mamba,dao2024transformers,liu2024vmamba} in control theory, which is based on the representation of continuous systems that model the input data with the ordinary differential equations (ODEs).
In contemporary SSMs, this continuous ODE is discretized. Mamba~\cite{gu2023mamba,dao2024transformers,liu2024vmamba} exemplifies such a discrete-time version of the continuous system, incorporating a timescale parameter $\boldsymbol{\Delta}$ to convert the continuous parameters $\mathbf{A}$ and $\mathbf{B}$ into their discrete equivalents $\overline{\mathbf{A}}$ and $\overline{\mathbf{B}}$ with the zero-order hold (ZOH) method as:

\begin{align}
\overline{\mathbf{A}} & = \exp(\boldsymbol{\Delta} \mathbf{A}) \label{eq:A} , \\
\overline{\mathbf{B}} & = (\boldsymbol{\Delta} \mathbf{A})^{-1}(\exp(\boldsymbol{\Delta} \mathbf{A}) - \mathbf{I}) \cdot \boldsymbol{\Delta} \mathbf{B} \label{eq:B} , \\
h_t & = \overline{\mathbf{A}} h_{t-1} + \overline{\mathbf{B}} x_t \label{eq:C} , \\
y_t & = \mathbf{C} h_t  \label{eq:D} .
\end{align}

Different from existing works that straightforwardly leverage mamba blocks to process input 2D or 3D patches~\cite{dao2024transformers,liu2024vmamba,liang2024pointmamba}, we propose to apply patch stacking and bidirectional mamba block for aligned contextual feature scanning. Specifically, we sequentially stack the depth-wise occupancy feature patches and the temporal-wise latent feature patches together to ensure that the most relevant features are scanned with relatively low contextual distance. 
Following that, the bidirectional mamba block is employed with simultaneous forward
and backward SSMs for spatially-aware enhancement. 
The output of the bidirectional mamba block $\bm{\widehat{X}}_{mam}$ is aggregated with the initial $\bm{L}$ through residual connection and zero convolution as ControlNet~\cite{zhang2023adding} to retain the inherent capabilities of the Diffusion UNet:

\begin{equation}
\bm{L}_{agg}= \bm{L} + \operatorname{Zero\_Conv}(\bm{\widehat{X}}_{mam}) .
\end{equation}

In this way, the Sequential Feature Encoding module integrates the semantic occupancy with the latent features with relevant contextual information. 

\begin{figure}[!t]
\begin{center}
\includegraphics[width=0.95\linewidth]{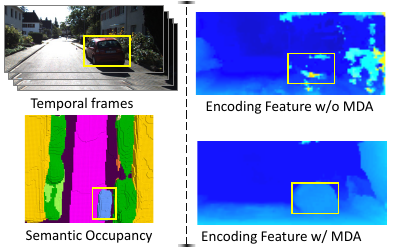}
 \vspace{-0pt}
\caption{Visualization results of the heat maps from our proposed Mamba-based Dual Alignment (MDA) module. The heat maps are extracted from the last diffusion sampling step. Our proposed module effectively highlights the aligned contextual information from the temporal frames and semantic occupancy.}
\label{learning}
\end{center}
 \vspace{-20pt}
\end{figure}

As depicted in Figure~\ref{learning}, the Mamba-based Dual Alignment (MDA) module effectively highlights the aligned contextual information from the temporal frames and semantic occupancy, while removing this module leads to blurred feature representation. 
Note that we implement cross-attention without Cross-view Camera Encoding between the temporal frames and semantic occupancy for the setting of `w/o MDA'.
For more details of other architecture design options and performance comparison on the MDA module, please refer to Section~\ref{sec_compare}.

\begin{figure}[!t]
\rcap
\begin{center}
\includegraphics[width=0.85\linewidth]{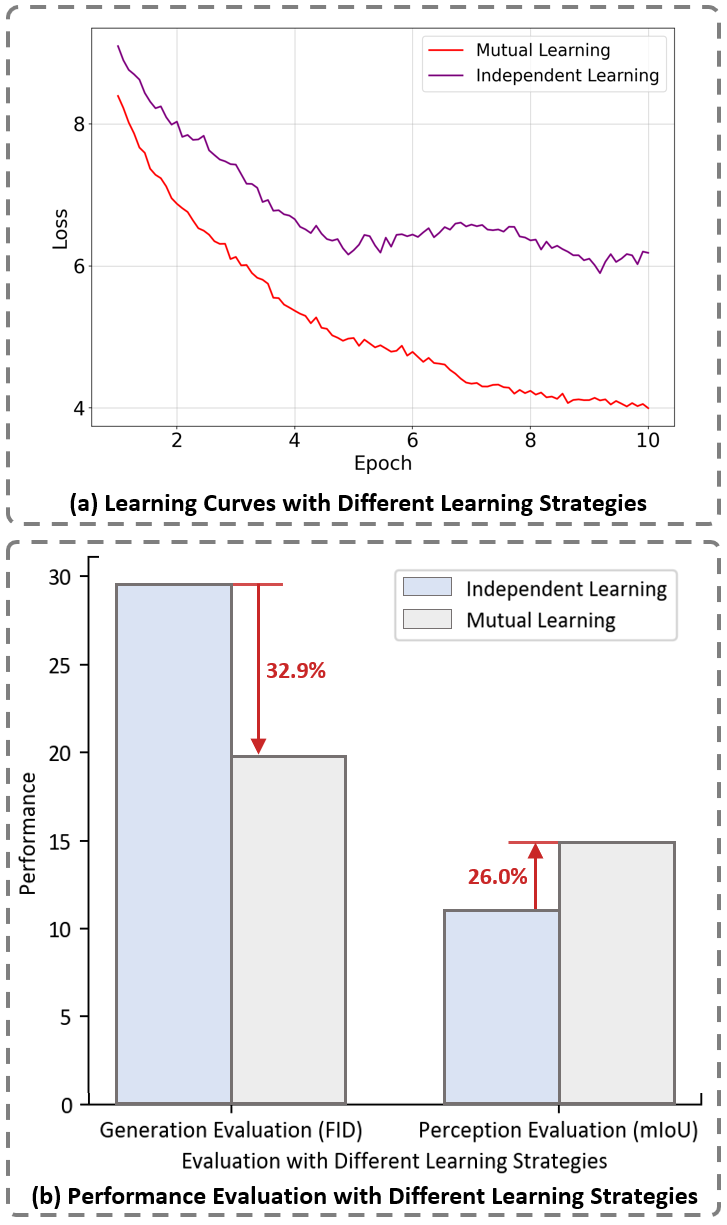}
\vspace{-0pt}
\caption{ 
\re{(a) The performance evaluation with different learning strategies. (b) The learning curves of different learning strategies. To conduct the setting of `Independent Learning', we
detach the perception model and predict the semantic occupancy in an offline manner.}}
\label{learning3}
\end{center}
 \vspace{-0pt}
\end{figure}

\subsection{\bh{Analysis on the Benefits of Cross-task Mutual Learning}}\label{sec_ana}

In this section, we analyze the mechanisms and rationale that underpin the effectiveness of our cross-task mutual learning strategy. Previous research on the generalization ability of deep neural networks has provided valuable insights~\cite{keskar2017large,chaudhari2019entropy,zhang2021understanding}. Notably, it has been found that among the many solutions (parameter configurations) that can achieve low training error, those with superior generalization tend to be located in wider valleys rather than narrower crevices of the loss landscape ~\cite{chaudhari2019entropy,zhang2021understanding}. These qualitative features are observed consistently across various network architectures, sizes, datasets, and optimization algorithms. Solutions identified by gradient descent that generalize well are typically situated in wide valleys, as opposed to sharp, isolated minima, which makes them more resilient to small perturbations without a significant impact on prediction accuracy. It is also noted that deep networks are particularly effective at finding these favorable solutions~\cite{chaudhari2019entropy}.\looseness=-1

Leveraging these insights, we observe that cross-task mutual learning facilitates the discovery of higher-quality solutions characterized by more robust minima. \re{As depicted in Figure~\ref{learning3}, we conducted extensive experiments to evaluate the impact of mutual learning. Our findings indicate that the framework performs better on training data when cross-task mutual learning is employed, leading to a more stable learning process and a better minimum of the training loss, as shown in Figure~\ref{learning3}(a). Specifically, while both independent and mutual learning initially reduce the loss, the independent learning strategy tends to stagnate in local minima during the middle stages of training, limiting its ability to converge effectively. In contrast, the mutual learning curve exhibits a steady decline throughout training, suggesting the identification of a broader and more optimal minimum, which indicates improved performance~\cite{chaudhari2019entropy,zhang2021understanding}.}
Figure~\ref{learning3}(b) further illustrates that through the joint learning scheme, our framework enables cross-task collaboration, leading to enhanced performance in both generation and perception tasks.

\section{Experiment}\label{sec_exp}

\subsection{Experimental Setup}
Our OccScene is implemented with PyTorch and trained with 8 NVIDIA A100 GPUs. 
For the generative backbone, we leverage pre-trained weights from SD~\cite{rombach2022high}.
Throughout the training, we freeze the SD model weights and only train the newly added parameters.
For the perception model, we leverage MonoScene~\cite{cao2022monoscene} with pre-trained weights and jointly train it with the generative framework.

\subsection{Datasets and Evaluation Metrics}\label{data}

\noindent\textbf{NYUv2.}
The NYUv2 dataset~\cite{silberman2012indoor} comprises 1449 indoor scenes captured via Kinect, represented as 240$\times$144$\times$240 occupancy grids annotated with 13 distinct classes: 11 semantic categories, alongside labels for free space and unknown areas. The accompanying RGBD input has a resolution of 640×480 pixels. 
Following~\cite{cao2022monoscene}, we employ 795 instances for training and 654 instances for testing.

\noindent\textbf{SemanticKITTI.}
The SemanticKITTI dataset~\cite{behley2019semantickitti} includes 22 diverse outdoor scenes featuring both LiDAR scans and stereo image pairs. Its ground truth is structured into 256$\times$256$\times$32 occupancy grids, each measuring 0.2m in all dimensions and annotated with 21 semantic categories, including 19 specific semantics, one class for free space, and another for unknown areas.\looseness=-1

\noindent\textbf{NuScenes-Occupancy.}
The nuScenes~\cite{caesar2020nuscenes} dataset is a prevalent autonomous driving dataset.
To enrich the dataset with fine-grained annotations, the nuScenes-Occupancy benchmark~\cite{wang2023openoccupancy} expanded it with dense semantic occupancy labels.
The benchmark encompasses 850 scenes, amounting to 34,000 keyframes with comprehensive LiDAR sweep data, each annotated with 17 semantic labels. 
Following~\cite{wang2023openoccupancy}, we allocate 28,130 frames for training and 6,019 for validation. 

\noindent\textbf{Metrics.}
For the evaluation metrics of perception results,
we adopt Mean Intersection over Union (mIoU) as the primary metric for evaluating the performance in semantic occupancy prediction (SOP) tasks following previous studies~\cite{cao2022monoscene,li2024time}. 
To evaluate generative results, we report the Frechet Inception Distance (FID)~\cite{heusel2017gans} and FVD (Frechet Video Distance)~\cite{unterthiner2018towards} 
scores to measure the generation quality.

\begin{table*}[!ht]
\subfloat[
\small{Generation quality on NYUv2 and SemanticKITTI}. 
]{\begin{minipage}{0.55\linewidth}
{
\small
\renewcommand\tabcolsep{4.6pt}
\resizebox{0.99\linewidth}{!}{
\begin{tabular}{l|ccccc}
    \toprule
    \multirow{2}{*}{Method}  &  \multicolumn{2}{c}{NYUv2} & \multicolumn{3}{c}{SemanticKITTI}  \\  \cmidrule{2-6} 
    & Resolution & FID$\downarrow$ & Resolution & FID$\downarrow$ & FVD$\downarrow$ \\ \midrule
    ControlNet~\cite{wang2023drivedreamer} & 448 $\times$ 640 & $50.61$  & 192 $\times$ 512 & $65.24$ & -\\
    SD (Finetune)~\cite{rombach2022high} & 448 $\times$ 640 & $47.82$  & 192 $\times$ 512 & $60.55$ & -   \\
    Tune-a-video~\cite{wu2023tune} & 448 $\times$ 640 &  -  & 192 $\times$ 512 & $\underline{55.93}$ & $\underline{209.41}$\\
    OccScene (ours) & 448 $\times$ 640 & $\textbf{15.54}$  & 192 $\times$ 512 & $\textbf{19.86}$ & $\textbf{113.28}$\\
    \bottomrule
  \end{tabular}
}
}
\end{minipage}
}
\hfill
\subfloat[
\small{Generation quality on NuScenes-Occupancy}. 
]{
\begin{minipage}{0.4\linewidth}
{
\small 
\renewcommand\tabcolsep{10.6pt}
\resizebox{0.99\linewidth}{!}{
\begin{tabular}{l|ccc}
    \toprule
    Method & Resolution & FID$\downarrow$  \\
    \midrule
    DriveGAN~\cite{kim2021drivegan} & 224 $\times$ 400 & 73.40  \\
    DriveDreamer~\cite{wang2023drivedreamer}  & 224 $\times$ 400 & 52.60 \\
    BEVGen~\cite{swerdlow2024streetview}  & 224 $\times$ 400 & 25.54  \\
    BEVControl~\cite{yang2023bevcontrol} & 224 $\times$ 400 &24.85  \\
    MagicDrive~\cite{gao2023magicdrive} & 224 $\times$ 400 &\underline{16.20}  \\
    OccScene (ours) & 256 $\times$ 448 & \textbf{11.87} \\
    \bottomrule
  \end{tabular}
  }
}

\end{minipage}
}
\vspace{-2pt}
\caption{{Comparison of generation fidelity on the NYUv2 test set, SemanticKITTI validation set and NuScenes-Occupancy validation set. The top two performers are marked \textbf{bold} and \underline{underline}. Our proposed method outperforms other methods in terms of image and video generation quality with equal or exceeding resolution.} }
\label{table:video_quality}
\end{table*}

\begin{figure*}[!ht]
\vspace{-0pt}
\hsize=\textwidth %
\centering
\includegraphics[width=0.9\textwidth]{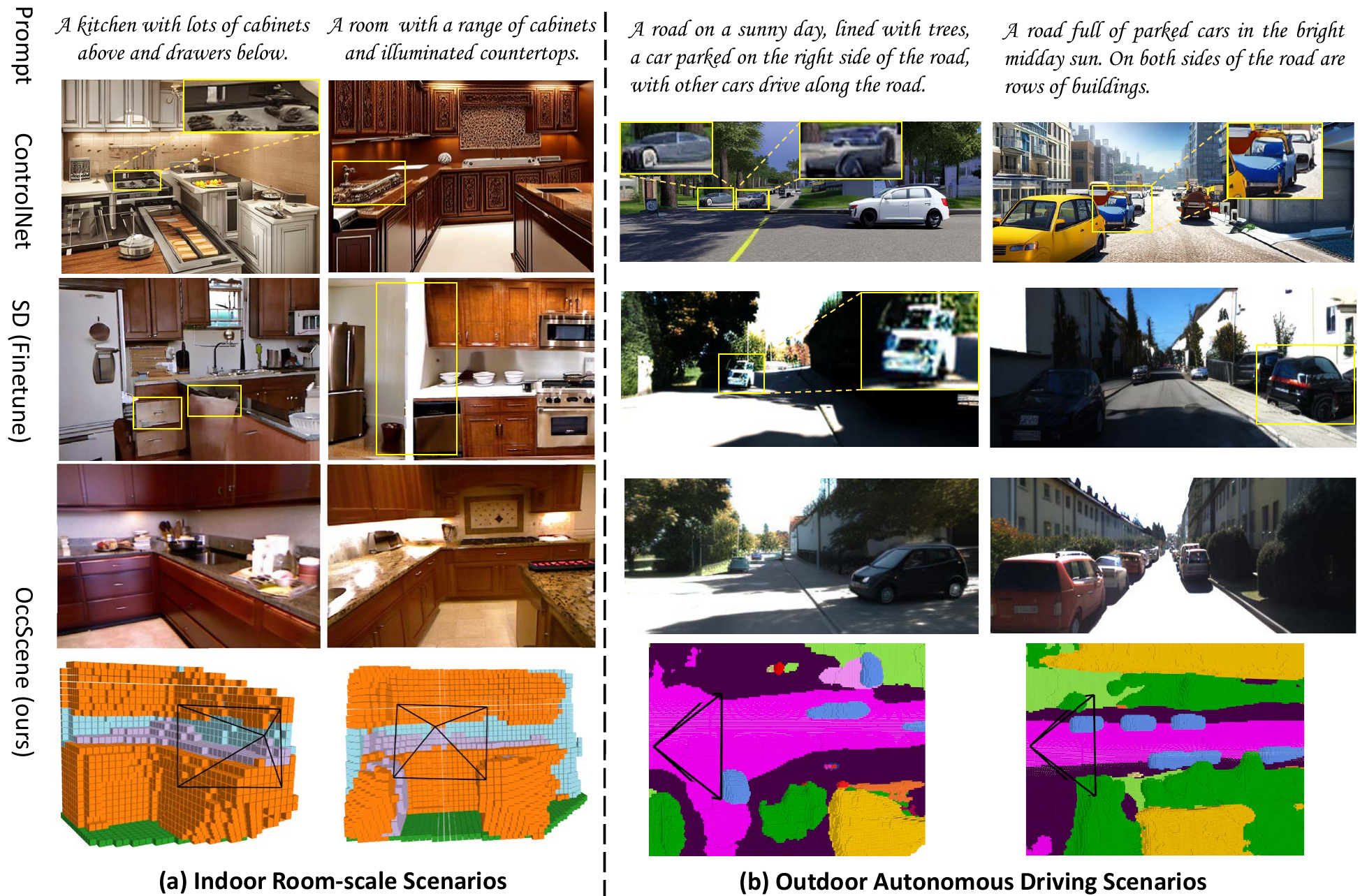}
\vspace{-0pt}
\caption{Quantitative comparison of scene generation with existing methods. The compared methods of SD~\cite{rombach2022high} and ControlNet~\cite{zhang2023adding} tend to generate unreasonable geometry and blurred details, especially in complex scenes and distant regions.}
\label{vis1}
\vspace{-0pt}
\end{figure*}

\begin{figure*}[!t]
\vspace{-0pt}
\hsize=\textwidth %
\centering
\includegraphics[width=0.99\textwidth]{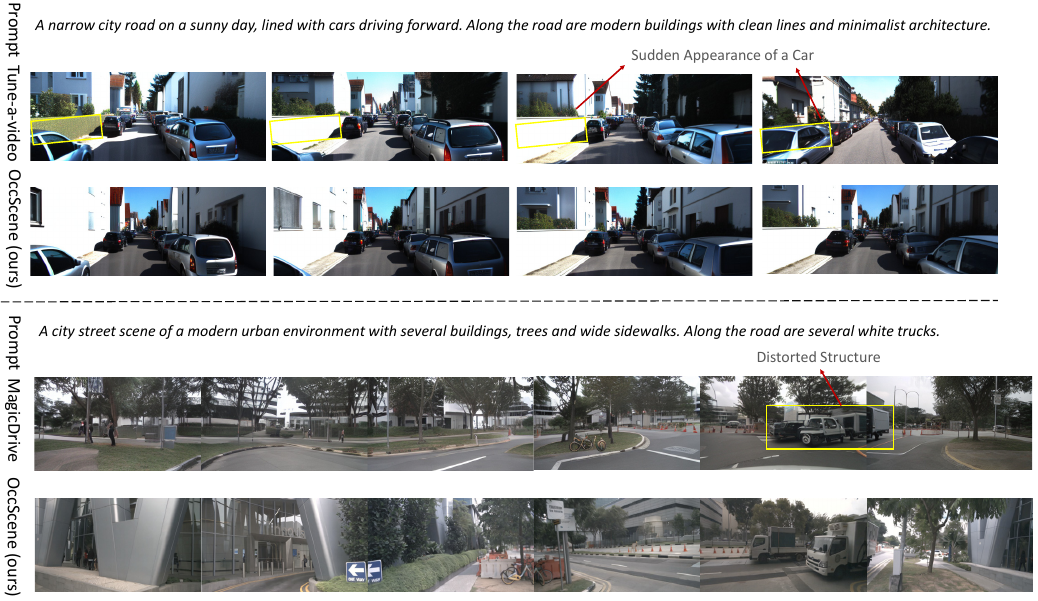}
\vspace{-0pt}
\caption{Quantitative comparison of cross-view generation consistency with existing methods. Our method generates more consistent and reasonable results across different perspectives. }
\label{vis2}
\vspace{-0pt}
\end{figure*}

\begin{table*}[!h]
\rcap
\begin{center}
\vspace{-0pt}
\renewcommand\tabcolsep{8.1pt}
\resizebox{0.99\textwidth}{!}{
		\begin{tabular}{l|c c c c c c c c c c c|c}
			\toprule
			Method
			& \rotatebox{90}{\textcolor{ceiling}{$\blacksquare$}ceiling}
			& \rotatebox{90}{\textcolor{floor}{$\blacksquare$} floor}
			& \rotatebox{90}{\textcolor{wall}{$\blacksquare$} wall} 
			& \rotatebox{90}{\textcolor{window}{$\blacksquare$} window} 
			& \rotatebox{90}{\textcolor{chair}{$\blacksquare$} chair} 
			& \rotatebox{90}{\textcolor{bed}{$\blacksquare$} bed} 
			& \rotatebox{90}{\textcolor{sofa}{$\blacksquare$} sofa} 
			& \rotatebox{90}{\textcolor{table}{$\blacksquare$} table} 
			& \rotatebox{90}{\textcolor{tvs}{$\blacksquare$} tvs} 
			& \rotatebox{90}{\textcolor{furniture}{$\blacksquare$} furniture} 
			& \rotatebox{90}{\textcolor{objects}{$\blacksquare$} objects} 
			& mIoU\\
			\midrule
			LMSCNet$^\text{rgb}$~\cite{roldao2020lmscnet}  & 4.49 & 88.41 & 4.63 & 0.25 & 3.94 & 32.03 & 15.44 & 6.57 & 0.02 & 14.51 & 4.39 & 15.88 \\
			AICNet$^\text{rgb}$~\cite{li2020anisotropic}  & 7.58 & 82.97 & 9.15 & 0.05 & 6.93 & 35.87 & 22.92 & 11.11 & 0.71 & 15.90 & 6.45 & 18.15 \\
			3DSketch$^\text{rgb}$~\cite{chen20203d}  & {8.53} & {90.45} &	{9.94} & {5.67} & {10.64} & {42.29} & {29.21} & {13.88} & {9.38} & {23.83} & {8.19} & {22.91}\\
		  MonoScene~\cite{cao2022monoscene}  & {8.89} & {93.50} & {12.06} & {12.57} & {13.72} & {48.19} & {36.11} & {15.13} & {15.22} & {27.96} & {12.94} & {26.94}\\

 NDC-Scene~\cite{Yao_2023_ICCV}  & 12.02 &\underline{93.51} &13.11& 13.77 & 15.83 & 49.57 & 39.87& 17.17& 24.57 &31.00 &14.96 &  29.03 \\
 
ISO~\cite{yu2024monocular} & \underline{14.21} & 93.47& \underline{15.89} & 15.14& \textbf{18.35} & \underline{50.01} & \underline{40.82} & 18.25& \underline{25.90} & \underline{34.08} & \underline{17.67} & \underline{31.25}
\\
\midrule
    \rowcolor{gray!10}MonoScene\cite{cao2022monoscene}$+$ours  &  {10.77} &  {93.48} & {15.72} &  \textbf{17.74} &   {15.76} &   {48.44} &   {40.33} & \underline{18.45} &   {17.33}&  {32.39} &   {17.14} & {29.78}  \\

\rowcolor{gray!10}NDC-Scene~\cite{Yao_2023_ICCV}$+$ours  &\textbf{15.06} &\textbf{94.85} &\textbf{16.51} &\underline{16.97} &\underline{17.84} &\textbf{51.10} &\textbf{42.41} &\textbf{19.56} &\textbf{26.33} &\textbf{34.51} &\textbf{18.14} & \textbf{32.12} \\
    
    \bottomrule
    \end{tabular}
    }
    \caption{ 
    \re{Quantitative results on the NYUv2 test set. The RGB-based variations of LMSCNet$^\text{rgb}$, AICNet$^\text{rgb}$, and 3DSketch$^\text{rgb}$ are implemented with RGB images as inputs. The top two performers are marked \textbf{bold} and \underline{underline}.}
    }
\vspace{-10pt}
\label{tabq1}
\end{center}
\end{table*}

\begin{table*}[!h]
\rcap
\begin{center}
\vspace{-10pt}
\renewcommand\tabcolsep{2.1pt}
\resizebox{0.99\textwidth}{!}{
\begin{tabular}{l|c c c c c c c c c c c c c c c c c c c |c}
    \toprule
    Method  
    & \rotatebox{90}{\textcolor{road}{$\blacksquare$}road} 
    & \rotatebox{90}{\textcolor{sidewalk}{$\blacksquare$} sidewalk} 
    & \rotatebox{90}{\textcolor{parking}{$\blacksquare$} parking}
    & \rotatebox{90}{\textcolor{other-ground}{$\blacksquare$} other-grnd}
    & \rotatebox{90}{\textcolor{building}{$\blacksquare$} building}
    & \rotatebox{90}{\textcolor{car}{$\blacksquare$} car}
    & \rotatebox{90}{\textcolor{truck}{$\blacksquare$} truck}
    & \rotatebox{90}{\textcolor{bicycle}{$\blacksquare$} bicycle}
    & \rotatebox{90}{\textcolor{motorcycle}{$\blacksquare$} motorcycle}
    & \rotatebox{90}{\textcolor{other-vehicle}{$\blacksquare$} other-veh.}
    & \rotatebox{90}{\textcolor{vegetation}{$\blacksquare$} vegetation}
    & \rotatebox{90}{\textcolor{trunk}{$\blacksquare$} trunk}
    & \rotatebox{90}{\textcolor{terrain}{$\blacksquare$} terrain}
    & \rotatebox{90}{\textcolor{person}{$\blacksquare$} person}
    & \rotatebox{90}{\textcolor{bicyclist}{$\blacksquare$} bicyclist}
    & \rotatebox{90}{\textcolor{motorcyclist}{$\blacksquare$} motorcyclist.}
    & \rotatebox{90}{\textcolor{fence}{$\blacksquare$} fence}
    & \rotatebox{90}{\textcolor{pole}{$\blacksquare$} pole}
    & \rotatebox{90}{\textcolor{traffic-sign}{$\blacksquare$} traf.-sign}  & mIoU
    \\
    \midrule
    
        LMSCNet$^\text{rgb}$~\cite{roldao2020lmscnet}   & 40.68 & 18.22 & 4.38 & 0.00 & 10.31 & 18.33 & 0.00 & 0.00 & 0.00 & 0.00 & 13.66 & 0.02 & 20.54 & 0.00 & 0.00 & 0.00 & 1.21 & 0.00 & 0.00 & 6.70 \\ %
      
      3DSketch$^\text{rgb}$~\cite{chen20203d}  & 41.32 & 21.63 & 0.00 & 0.00 & {14.81} & 18.59 & 0.00 & 0.00 & 0.00 & 0.00 & {19.09} & 0.00 & 26.40 & 0.00 & 0.00 & 0.00 & 0.73 & 0.00 & 0.00 & 7.50  \\ %

    AICNet$^\text{rgb}$~\cite{li2020anisotropic}  & 43.55 & 20.55 & {11.97} & {0.07} & 12.94 & 14.71 & {4.53} & 0.00 & 0.00 & 0.00 & 15.37 & {2.90} & {28.71} & 0.00 & 0.00 & 0.00 & 2.52 & 0.06 & 0.00 & 8.31  \\ 
      
    MonoScene~\cite{cao2022monoscene}  & 56.52 &26.72 &14.27 &0.46 &14.09 &23.26 &6.98 &0.61 &0.45 &1.48 &17.89 &2.81 &29.64 &1.86 &1.20 &0.00 &5.84 &4.14 &2.25  &11.08  \\

    TPVFormer~\cite{huang2023tri} & 56.50  & 25.87 & {20.60}  & {0.85} & 13.88 & 23.81  & {8.08}  & 0.36  & 0.05   & {4.35}  & 16.92  & 2.26   & 30.38  & 0.51  & 0.89 & 0.00  & 5.94   & 3.14 & 1.52  & 11.36\\
    VoxFormer~\cite{li2023voxformer}   & 54.76  & 26.35 & 15.50  & 0.70  & {17.65}  & {25.79} & 5.63   & {0.59}   & 0.51  & 3.77   & {24.39}   & {5.08}  & 29.96   & 1.78  &{3.32} & 0.00  & {7.64}   & {7.11}  & {4.18}  & {12.35}  \\  
    OccFormer~\cite{zhang2023occformer}  & 58.85  & 26.88 & 19.61 & 0.31 & 14.40 & 25.09 & \underline{25.53} & 0.81  & 1.19  & {8.52} & 19.63  & 3.93  & {32.62} & 2.78 & 2.82  & 0.00  & 5.61  & 4.26  & 2.86 & 13.46         \\
    CGFormer~\cite{yu2024context}  &  \underline{65.51} & 32.31& 20.82& 0.16& \underline{23.52} & \underline{34.32} & {19.44} & \textbf{4.61} & \underline{2.71} & 7.67& \underline{26.93} & \underline{8.83} & \underline{39.54} & 2.38& \underline{4.08} & 0.00& 9.20& 10.67& \textbf{7.84} &  \underline{16.87}   \\  \midrule
    
    \rowcolor{gray!10}MonoScene\cite{cao2022monoscene}$+$ours  & {62.59}  & \underline{33.20} & \textbf{22.65} & \textbf{3.41} & {19.40} & {26.67} & {14.27} &  {1.85} & {2.07} & {7.00} &{22.54} &{5.11} &\textbf{39.95} &\textbf{4.42} &{1.46} &{0.00} &{8.08} &{6.42} &{3.49} &{14.98}\\  

    \rowcolor{gray!10}TPVFormer~\cite{huang2023tri}$+$ours & {61.93} &{32.95} &\underline{20.89} &0.21 &{23.24} &{32.46} &{15.97} &{2.41} &{1.98} &\underline{8.82} &{26.20} &{8.04} &{33.27} &{2.52} &0.87 &0.00  &\underline{9.39} &\underline{10.90} &\underline{6.80} &{15.73} \\ 
    
    \rowcolor{gray!10}OccFormer~\cite{zhang2023occformer}$+$ours   & \textbf{65.73} &\textbf{33.96} &20.22 &\underline{1.20} &\textbf{23.65} &\textbf{34.85} &\textbf{26.58} &\underline{3.37} &\textbf{2.79} &\textbf{10.63} &\textbf{27.85} &\textbf{8.97} &36.86 &\underline{3.00} &\textbf{4.22} &0.00 &\textbf{10.36} &\textbf{11.25} &5.74 & \textbf{17.43}
     \\

    \bottomrule
    \end{tabular}
    }
\caption{ 
\re{Quantitative results on the SemanticKITTI validation set. The RGB-based variations of LMSCNet$^\text{rgb}$, AICNet$^\text{rgb}$, and 3DSketch$^\text{rgb}$ are implemented with RGB images as inputs. The top two performers are marked \textbf{bold} and \underline{underline}.}
}
\vspace{-10pt}
\label{tabq2}
\end{center}
\end{table*}

\begin{table*}[!t]
\renewcommand\tabcolsep{5.6pt}
	\centering
   \resizebox{0.99\textwidth}{!}{
	\begin{tabular}{l| c  c c c c c c c c c c c c c c c | c}
 
		\toprule
		Method
		& \rotatebox{90}{\textcolor{barrier}{$\blacksquare$} barrier} 
		& \rotatebox{90}{\textcolor{bicycle}{$\blacksquare$} bicycle}
		& \rotatebox{90}{\textcolor{bus}{$\blacksquare$} bus} 
		& \rotatebox{90}{\textcolor{car}{$\blacksquare$} car} 
		& \rotatebox{90}{\textcolor{const. veh.}{$\blacksquare$} const. veh.} 
		& \rotatebox{90}{\textcolor{motorcycle}{$\blacksquare$} motorcycle} 
		& \rotatebox{90}{\textcolor{pedestrian}{$\blacksquare$} pedestrian} 
		& \rotatebox{90}{\textcolor{traffic cone}{$\blacksquare$} traffic cone} 
		& \rotatebox{90}{\textcolor{trailer}{$\blacksquare$} trailer} 
		& \rotatebox{90}{\textcolor{truck}{$\blacksquare$} truck} 
		& \rotatebox{90}{\textcolor{drive. suf.}{$\blacksquare$} drive. suf.} 
		& \rotatebox{90}{\textcolor{other flat}{$\blacksquare$} other flat} 
		& \rotatebox{90}{\textcolor{sidewalk}{$\blacksquare$} sidewalk} 
		& \rotatebox{90}{\textcolor{terrain}{$\blacksquare$} terrain} 
		& \rotatebox{90}{\textcolor{manmade}{$\blacksquare$} manmade} 
		& \rotatebox{90}{\textcolor{vegetation}{$\blacksquare$} vegetation} & \makecell[c]{mIoU} \\
		\midrule
		MonoScene~\cite{cao2022monoscene}   & 7.1  & 3.9  &  9.3 &  7.2 & 5.6  & 3.0  &  5.9& 4.4& 4.9 & 4.2 & 14.9 & 6.3  & 7.9 & 7.4  & 10.0 & 7.6 & 6.9\\
  
  		TPVFormer~\cite{huang2023tri}   & 9.3  & 4.1  &  11.3 &  10.1 & 5.2  & 4.3  & 5.9 & 5.3&  6.8& 6.5 & 13.6 & 9.0  & 8.3 & 8.0  & 9.2 & 8.2 &  7.8 \\ \midrule
            
            AICNet$^\text{*}$~\cite{li2020anisotropic}  & \underline{11.5}  & 4.0  & \underline{11.8}  & 12.3&  5.1 & 3.8  & 6.2  & \underline{6.0} & \textbf{8.2} &  7.5&  24.1 & 13.0 & 12.8  & 11.5 & \underline{11.6}  &  \underline{20.2} & 10.6 \\ 
            
            3DSketch$^\text{*}$~\cite{chen20203d}  & \textbf{12.0} &  5.1 &  10.7 &  12.4 & \underline{6.5}  & 4.0  & 5.0 & \textbf{6.3} &  \underline{8.0}&  7.2& 21.8 &  14.8 & 13.0 &  11.8 & \textbf{12.0} & \textbf{21.2}  & 10.7 \\ \midrule

\rowcolor{gray!10}  MonoScene\cite{cao2022monoscene}$+$ours & 9.7& \underline{6.4} & 11.0& \underline{12.6}& 6.3& \underline{7.1} & \underline{8.3} & 4.4& 7.3& \underline{10.7} & \underline{24.8} & \underline{15.1}& \textbf{17.1} & \underline{14.7} & 8.8& 11.8 & \underline{11.0}  \\ 

\rowcolor{gray!10}    TPVFormer~\cite{huang2023tri}{$+$}ours  & \underline{10.4}&  \textbf{7.3}&  \textbf{12.4}&  \textbf{13.5} &  \textbf{9.6}& \textbf{10.9} & \textbf{10.7}& 5.0& 7.6& \textbf{11.8} & \textbf{24.9} & \textbf{17.5} & \underline{16.8}& \textbf{15.2} & 9.3& 12.6 &\textbf{12.2}\\ 
		\bottomrule
	\end{tabular}}
    \caption{Quantitative results on the NuScenes-Occupancy validation set. The AICNet$^\text{*}$ and 3DSketch$^\text{*}$ take images and LiDAR-projected depth maps as inputs. The top two performers are marked \textbf{bold} and \underline{underline}.}
    
	\label{table_nuscene}
\end{table*}

\noindent\subsection{Main Results}

\noindent\textbf{Generation Evaluation.} 
The quantitative results of scene generation are presented in Figure~\ref{vis1}.
For the baseline models of SD~\cite{rombach2022high}, we fine-tune it on the datasets with the same training setting as OccScene. ControlNet~\cite{wang2023drivedreamer} is conditioned with semantic maps and depth maps jointly to generate corresponding images.
As illustrated in Figure~\ref{vis1}, both SD and ControlNet tend to generate unreasonable geometry (e.g., cars in columns 3 and 4) and blurred details (e.g., distant structures in columns 1 and 2), especially in complex scenes and distant regions. 
We also present the quantitative results of cross-view generation in Figure~\ref{vis2}. The model of tune-a-video~\cite{wu2023tune} is further fine-tuned on the SemanticKITTI dataset. Compared to existing models, our method generates more consistent and reasonable results across different perspectives.
The significant superiority stems from the incorporated fine-grained geometry and semantics as the perception priors. 
Moreover, we report the quantitative results with different datasets in Table~\ref{table:video_quality}(a) and Table~\ref{table:video_quality}(b). Our proposed OccScene outperforms other methods in terms of image and video generation
with equals or exceeds baseline resolution of 256 $\times$ 448, achieving 113.28 FVD on SemanticKITT~\cite{behley2019semantickitti} and 11.87 FID on  NuScenes-Occupancy~\cite{wang2023openoccupancy}. The extensive experiments demonstrate the effectiveness of our proposed method on high-fidelity scene generation.\looseness=-1

\begin{table}[!h]
\rcap
\vspace{-0pt}
\centering
\renewcommand\tabcolsep{16.0pt}
\resizebox{0.99\columnwidth}{!}{
\begin{tabular}{l|ccc}
\toprule
Method & FID$\downarrow$ & KID$\downarrow$ & IS$\uparrow$ \\ \midrule
    SSD~\cite{lee2023diffusion}    & 112.82 & 0.12 & 2.23  \\ 
    SemCity~\cite{lee2024semcity}   & \underline{56.55} & \underline{0.04} & \underline{3.25}  \\ 
   \rowcolor{gray!10} OccScene (Ours)  & \textbf{39.21} & \textbf{0.02}  &\textbf{4.17} \\ \bottomrule
\end{tabular}
}
\caption{
\re{Quantitative comparison of 3D semantic scene generation performance on the SemanticKITTI validation set. The top two performers are marked \textbf{bold} and \underline{underline}. Our proposed method outperforms previous works in terms of the 3D semantic scene generation quality.}} 
\label{table_occ_gen}
\vspace{-0pt}
\end{table}

\begin{figure*}[!h]
\rcap
\vspace{-0pt}
\hsize=\textwidth %
\centering
\includegraphics[width=0.81\textwidth]{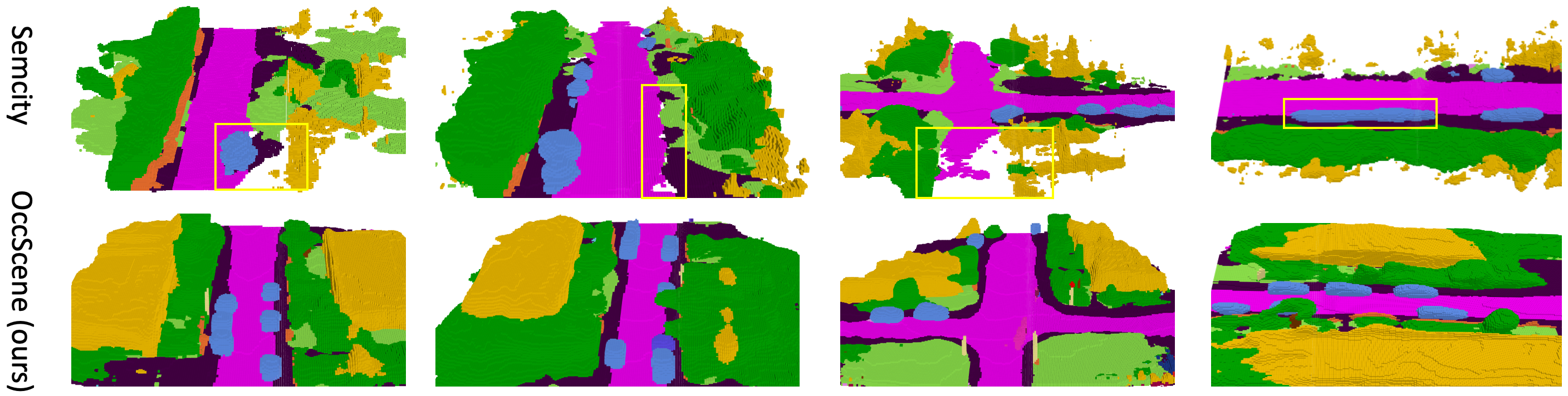}
\vspace{-0pt}
\caption{
\re{Quantitative comparison of 3D semantic scene generation with SemCity~\cite{lee2024semcity}. Our method efficiently generates more complete scenes with detailed structures, especially in road surfaces and vehicle shapes.}
}
\label{figure_occ_compare}
\vspace{-0pt}
\end{figure*}

\re{The qualitative evaluation results of 3D semantic scene generation are shown in Table~\ref{table_occ_gen}. Following SemCity~\cite{lee2024semcity}, we assess generation quality using Frechet Inception Distance (FID)~\cite{heusel2017gans}, Kernel Inception Distance (KID)~\cite{binkowski2018demystifying}, and Inception Score (IS). 
Our proposed OccScene outperforms previous works in terms of generation quality, 
achieving a 30.66\% improvement in FID compared to SemCity~\cite{lee2024semcity} and a 65.25\% improvement compared to SSD~\cite{lee2023diffusion}. These results demonstrate the effectiveness of our approach for high-fidelity 3D scene generation.}
\re{The quantitative evaluation results of 3D semantic scene generation are illustrated in Figure~\ref{figure_occ_compare}.
\emph{Note that we independently replicated the semantic scene generation model of SemCity~\cite{lee2024semcity} following their official implementations, as no pretrained weights are available in the public repository.} Facilitated by the cross-task mutual benefits, our OccScene produces more realistic and complete 3D scene generation results, especially in overall completion (\textit{e.g.}, road surfaces in the first three columns) and structural details (\textit{e.g.}, vehicle shapes in the fourth column).}

\noindent\textbf{Perception Evaluation.} 
We compare our method with other state-of-the-art SOP networks~\cite{roldao2020lmscnet,li2020anisotropic,cao2022monoscene,chen20203d,li2023voxformer,huang2023tri,Yao_2023_ICCV,yu2024monocular,zhang2023occformer,yu2024context} for perception evaluation on the NYUv2 dataset in Table~\ref{tabq1} and the SemanticKITTI dataset in Table~\ref{tabq2}.
Following MonoScene~\cite{cao2022monoscene}, we only adopt RGB images as inputs and implement RGB-based variations of LMSCNet$^\text{rgb}$~\cite{roldao2020lmscnet}, AICNet$^\text{rgb}$~\cite{li2020anisotropic} and 3DSketch$^\text{rgb}$~\cite{chen20203d}.
\re{To further demonstrate the effectiveness of OccScene as a general plug-and-play framework for enhancing downstream task performance, we conducted additional experiments using MonoScene~\cite{cao2022monoscene} and NDC-Scene~\cite{Yao_2023_ICCV} as baselines on the NYUv2 test set (see Table~\ref{tabq1}), and MonoScene~\cite{cao2022monoscene}, TPVFormer~\cite{huang2023tri}, and OccFormer~\cite{zhang2023occformer} on the SemanticKITTI validation set (see Table~\ref{tabq2}).
As shown in the tables, our method achieves significant improvements, increasing the mIoU by 2.84 for MonoScene~\cite{cao2022monoscene} and 3.09 for NDC-Scene~\cite{Yao_2023_ICCV} on the NYUv2 test set, and by 3.90 for MonoScene~\cite{cao2022monoscene}, 4.38 for TPVFormer~\cite{huang2023tri}, and 3.97 for OccFormer~\cite{zhang2023occformer} in semantic occupancy prediction. These results further validate OccScene's effectiveness in boosting downstream task performance.}
Moreover, we evaluate the effectiveness of OccScene on the OpenOccupancy validation set in Table~\ref{table_nuscene}.
Following~\cite{wang2023openoccupancy}, we only adopt RGB images as inputs for MonoScene~\cite{cao2022monoscene} and TPVFormer~\cite{huang2023tri}, while the AICNet$^\text{*}$~\cite{li2020anisotropic} and 3DSketch$^\text{*}$~\cite{chen20203d} take images and depth maps as inputs.
To provide depth maps for them, LiDAR points are projected and densified following OpenOccupancy~\cite{wang2023openoccupancy}.
We adopt MonoScene~\cite{cao2022monoscene} and TPVFormer~\cite{huang2023tri} as baseline models to highlight the superior capabilities of our OccScene to improve the performance in the downstream perception task.
As shown in the table, our method improves 4.10 mIoU for MonoScene and 4.40 mIoU for TPVFormer in semantic occupancy prediction, underscoring the efficacy of OccScene as a general plug-and-play framework in improving downstream task performance.\looseness=-1

\begin{figure}[!t]
 \vspace{-0pt}
\begin{center}
\includegraphics[width=0.99\linewidth]{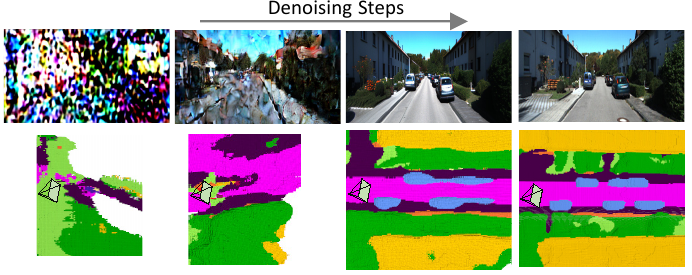}
 \vspace{-0pt}
\caption{Generation results of different denoising steps. As the generated images become clearer, the semantic occupancy predicted by the perception model becomes more complete and accurate.\looseness=-1}
\label{learning1}
\end{center}
\end{figure}

\begin{figure}[!h]
\rcap
\begin{center}
\includegraphics[width=0.85\linewidth]{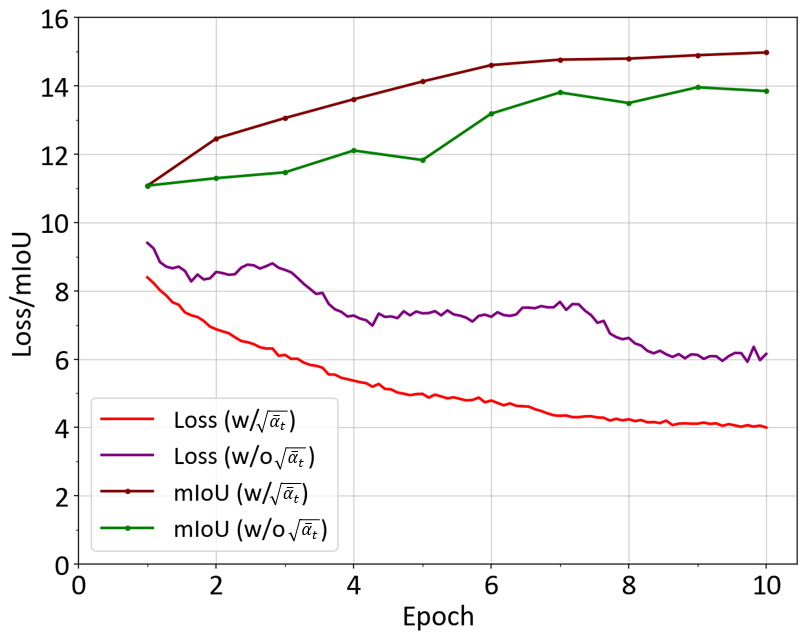}
 \vspace{-0pt}
\caption{ 
\re{The learning curves with $\sqrt{\bar{\alpha}_t}$. 
The stability of the loss curve and the perception performance with mIoU scores are significantly improved by applying $\sqrt{\bar{\alpha}_t}$.}
}
\label{learning2}
\end{center}
 \vspace{-0pt}
\end{figure}

\noindent\textbf{Learning Process Analysis.} 
The results generated from different denoising steps are visualized in Figure~\ref{learning1}. \re{As the generated images become clearer, the semantic occupancy predicted by the perception model becomes more complete and accurate.
As discussed in Section III.B, to stabilize training, the loss function incorporates $\sqrt{\bar{\alpha}_t}$ to adaptively scale supervision signals according to noise levels.
As shown in Figure~\ref{learning2}, this modification significantly improves the stability of the loss curve and enhances the perception performance, as evidenced by higher mIoU scores.}

\begin{table}[!t]
\vspace{-0pt}
\centering
\renewcommand\tabcolsep{10.1pt}
\resizebox{0.9\columnwidth}{!}{
\begin{tabular}{l|cc}
\toprule
Data  & IoU$\uparrow$  & mIoU$\uparrow$   \\ \midrule

w/o synthetic data & 18.4 &6.9   \\
 w/ MagicDrive & 17.8 ${}_{{-0.6}}$  & 7.2 ${}_{{+0.2}}$ \\
 w/  OccScene & \textbf{21.3} ${}_{{+2.9}}$  &\textbf{10.2} ${}_{{+3.3}}$ \\ \bottomrule
\end{tabular}
}
\caption{\bh{ Comparison about support for Semantic Occupancy Prediction model (i.e., MonoScene). Results are reported by testing on the NuScenes-Occupancy validation set.}.} 
\label{table_sup1}
\vspace{-0pt}
\end{table}

\noindent\textbf{Training Support for Semantic Occupancy Prediction.} 
\bh{As shown in Table~\ref{table_sup1}, we conduct a training support experiment to demonstrate that OccScene can generate synthetic image-occupancy data pairs to enhance the training for the perception task of Semantic Occupancy Prediction.
To produce the data pairs, we generate the same amount of images as the original dataset. Note that the Semantic Occupancy Prediction model of MonoScene~\cite{cao2022monoscene} is trained from scratch on the synthetic data to enable fair comparisons.
As shown in the table, OccScene significantly improves MonoScene in terms of both IoU and mIoU for Semantic Occupancy Prediction. While the compared method of MagicDrive~\cite{gao2023magicdrive} only
marginally improves mIoU. We attribute such a difference to the high fidelity and fine-grained geometric control of generated results from OccScene.}

\begin{table}[!t]
\vspace{-0pt}
\centering
\renewcommand\tabcolsep{12.0pt}
\resizebox{0.9\columnwidth}{!}{
\begin{tabular}{l|ccc}
\toprule
Component & FID$\downarrow$ & FVD$\downarrow$ & mIoU$\uparrow$ \\ \midrule

 w/o JDS &28.52 & 187.21 &12.94 \\
 w/o MDA &25.71  & 162.04 & 13.41  \\
 w/ JDS$\&$MDA & \textbf{19.86} & \textbf{113.28} & \textbf{14.98}  \\ \bottomrule
\end{tabular}
}
\caption{Ablation studies of the framework components on the SemanticKITTI validation set. The `JDS' and `MDA' denote the Joint Diffusion Scheme and the Mamba-based Dual Alignment, respectively.} 
\label{table_ar1}
\vspace{-0pt}
\end{table}

\subsection{Ablation Study}
To validate the effectiveness of our proposed framework components, we conduct extensive ablation studies on the SemanticKITTI validation set.

\noindent\textbf{Joint Diffusion Scheme (JDS).}
As shown in Table~\ref{table_ar1}, to conduct the setting of `w/o JDS' for effect evaluation, we detach the perception model and predict the semantic occupancy in an offline manner. Specifically, the generative framework is conditioned on the pre-produced occupancy to generate corresponding images, and the generated images of the last inference step are leveraged to train the perception model. As illustrated in the table, the joint diffusion scheme benefits the fidelity of image and video generation significantly.
Moreover, the joint scheme enhances 2.04 mIoU for the perception performance compared to the offline strategy, which stems from the utilization of different information capacities in the generation process.

\begin{table}[!h]
\rcap
\vspace{-0pt}
\centering
\renewcommand\tabcolsep{1.5pt}
\resizebox{0.99\columnwidth}{!}{
\begin{tabular}{l|cc}
\toprule
Method  & IoU$\uparrow$  & mIoU$\uparrow$   \\ \midrule

MonoScene~\cite{cao2022monoscene}  &42.51 & 26.94 \\
 
MonoScene~\cite{cao2022monoscene} + OccScene (detached gradients) & 43.05  & 27.47\\ 
  
\rowcolor{gray!10}  MonoScene~\cite{cao2022monoscene} + OccScene (attached gradients) & 44.34  &29.78 \\  \midrule

NDC-Scene~\cite{Yao_2023_ICCV} &44.17 & 29.03 \\
 
NDC-Scene~\cite{Yao_2023_ICCV} + OccScene (detached gradients) &    45.62 & 30.20  \\ 
  
\rowcolor{gray!10}  NDC-Scene~\cite{Yao_2023_ICCV} + OccScene (attached gradients) &  47.89 & 32.12 \\  \midrule

ISO~\cite{yu2024monocular} &47.11  & 31.25 \\
 
ISO~\cite{yu2024monocular} + OccScene (detached gradients) & 48.60 & 32.09   \\ 
  
\rowcolor{gray!10}  ISO~\cite{yu2024monocular} + OccScene (attached gradients) & 50.07 & 33.92  \\  \midrule

\end{tabular}
}
\caption{\re{Effect of the joint training methodology on the NYUv2 test set, which effectively improves extensive perception models.}} 
\label{table_joint_occ}
\vspace{-0pt}
\end{table}

\re{We evaluate the effect of the joint training methodology with different perception models on the NYUv2 test set, as shown in Table~\ref{table_joint_occ}. 
Specifically, the setting of `OccScene (detached gradients)' represents training baseline models (\textit{e.g.,} MonoScene~\cite{cao2022monoscene}, NDC-Scene~\cite{Yao_2023_ICCV}, ISO~\cite{yu2024monocular}) with gradients detached from the generative model but with identical data augmentation (including augmented noisy data).
The setting of `OccScene (attached gradients)' represents training baseline models with gradients attached to the generative model.
The `attached gradients' setting consistently outperforms the `detached gradients' setting across all baseline models, improving 3.72 IoU and 3.09 mIoU for NDC-Scene~\cite{Yao_2023_ICCV}. This performance enhancement underscores the effectiveness of our joint training methodology in improving extensive perception models.}\looseness=-1

\begin{figure*}[!ht]
\begin{center}
\vspace{-0pt}
\includegraphics[width=0.9\linewidth]{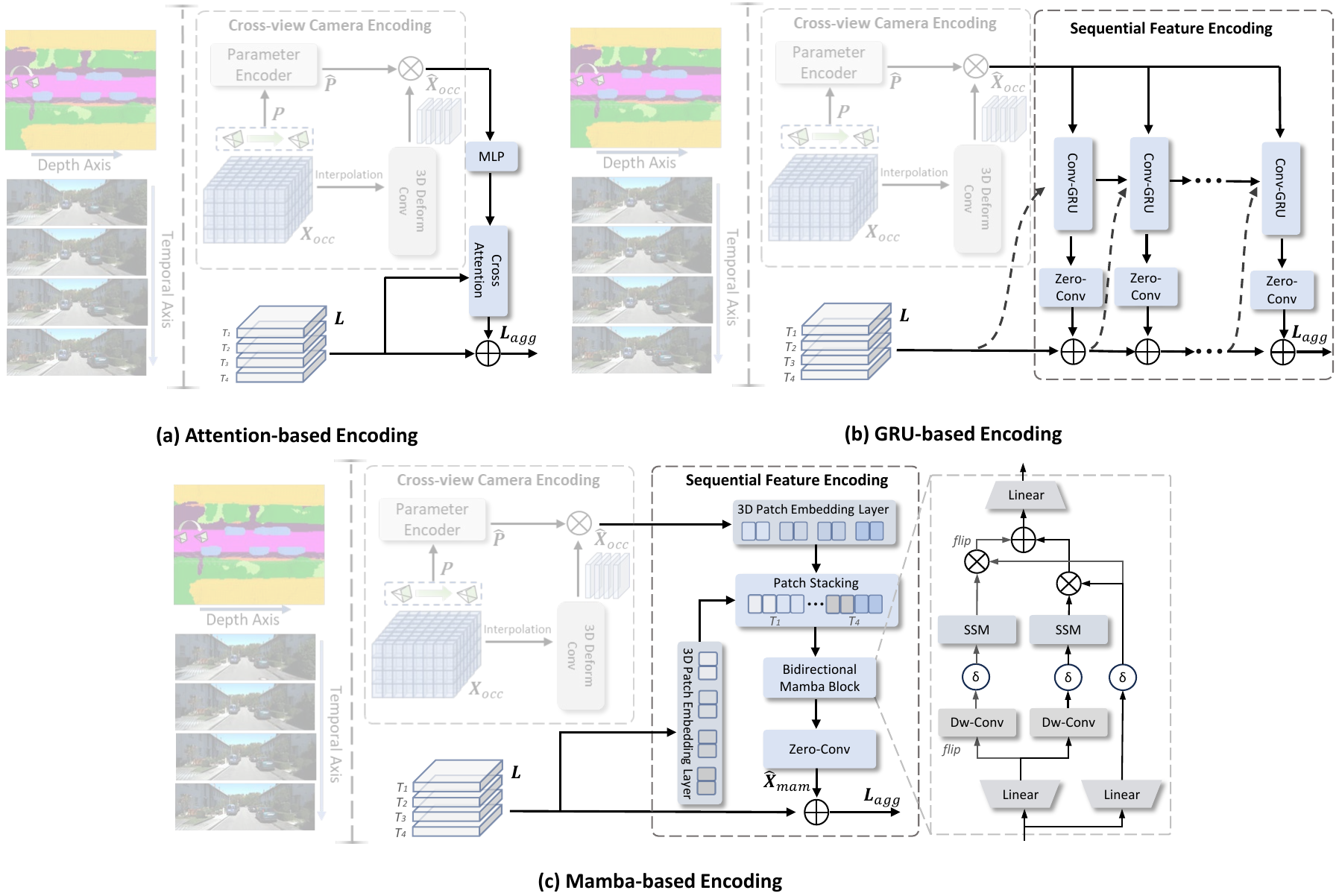}
 \vspace{-0pt}
\caption{Comparison of different architecture designs of the Mamba-based Dual Alignment (MDA) module.
Although the GRU-based encoding yields better computational efficiency through an iterative encoding process compared to the Attention-based encoding, it is susceptible to cumulative errors.
The Mamba-based encoding demonstrates superior running speed and generation quality with the linear-complexity operator and efficient long-term modeling in a single pass.
}
\label{encode}
\end{center}
 \vspace{-0pt}
\end{figure*}

\begin{table}[!t]
\vspace{-0pt}
\centering
\renewcommand\tabcolsep{10.5pt}
\resizebox{0.9\columnwidth}{!}{
\begin{tabular}{l|ccc}
\toprule
Architecture & FID$\downarrow$ & FVD$\downarrow$ & Time$\downarrow$ \\ \midrule
    Attention-based    & 25.71 & 162.04  &4.09  \\ 
    GRU-based   & 24.54 & 135.71  &3.27  \\ 
    Mamba-based   & \textbf{19.86} & \textbf{113.28}  &\textbf{2.76} \\ \bottomrule
\end{tabular}
}
\caption{\bh{Comparison of generation quality with different architecture designs of the Mamba-based Dual Alignment (MDA) module on the SemanticKITTI
validation set. }} 
\label{table_ar2}
\vspace{-0pt}
\end{table}

\noindent\textbf{Mamba-based Dual Alignment.}\label{sec_compare} 
As shown in Figure~\ref{encode}, to evaluate the effect of the MDA module, we design and compare different encoding architectures including Attention-based encoding, GRU-based encoding and Mamba-based encoding. 
Note that the setting of `w/o MDA' in Table~\ref{table_ar1} is conducted with cross-attention without Cross-view Camera Encoding.
Please refer to the supplementary material for architectural details of different designs.

Compared to attention-based encoding, the GRU-based architecture is a better choice to sequentially encode these high-dimension features through iterative processing for computational efficiency.
However, due to the inability to process all the input information in a single pass, such an approach is susceptible to cumulative errors~\cite{li2017diffusion, mao2022review}.

As shown in Table~\ref{table_ar2}, the mamba-based demonstrates superior running speed and generation quality, which we attribute to the linear-complexity operator and efficient long-term modeling to process input high-dimensional data in a single pass.
Specifically, the mamba-based architecture reduces 32.52\% running time compared to the attention-based design and 19.07\% FID compared to the GRU-based design, respectively.\looseness=-1

\begin{table}[!h]
\rcap
\vspace{-0pt}
\centering
\renewcommand\tabcolsep{12.0pt}
\resizebox{0.9\columnwidth}{!}{
\begin{tabular}{l|ccc}
\toprule
Component & FID$\downarrow$ & FVD$\downarrow$ & mIoU$\uparrow$ \\ \midrule
 w/o MDA-D & 22.12  & 121.84  & 14.36  \\
 w/o MDA-T & 23.02  & 128.17  & 14.04 \\
 w/ MDA & \textbf{19.86} & \textbf{113.28} & \textbf{14.98}  \\ \bottomrule
\end{tabular}
}
\caption{\re{Ablation study of the Mamba-based Dual Alignment (MDA) module applied along different dimensions on the SemanticKITTI validation set.  
}
} 
\label{table_ar_mamba}
\vspace{-0pt}
\end{table}

\re{Table~\ref{table_ar_mamba} illustrates the effect of applying the MDA module along different dimensions.
For the setting of ` w/o MDA-D', the bidirectional Mamba block is applied exclusively to the video diffusion latent $\bm{L} \in C_L \times N \times H_L \times W_L$, while 3D convolution layers are used for the semantic occupancy $\bm{{X}}_{occ} \in 1\times D\times H_{occ} \times W_{occ}$. Conversely, the setting of ` w/o MDA-D' employs the bidirectional Mamba block only for the semantic occupancy $\bm{{X}}_{occ}$, with 3D convolutions handling the video latent feature $\bm{L}$.
As shown in the table, applying the MDA module along both the depth and temporal dimensions yields substantial performance enhancements, improving FID by 2.26 and 3.16, respectively.}\looseness=-1

\begin{table}[!ht]
\vspace{-0pt}
\centering
\renewcommand\tabcolsep{3.1pt}
\resizebox{0.9\columnwidth}{!}{
\begin{tabular}{l|cccc}
\toprule
Dataset  & Resolution & Steps  & FID$\downarrow$  & Time$\downarrow$    \\ \midrule
\multirow{3}{*}{NYUv2}  & 448 $\times$ 640 & 20 &19.75 & 1.72\\   
  & 448 $\times$ 640 & 50 &15.54 & 3.74 \\ 
  & 448 $\times$ 640 & 100 & 14.34 & 7.35 \\ \midrule

\multirow{3}{*}{SemanticKITTI} & 192 $\times$ 512 & 20  &  22.93 & 1.64 \\  
 & 192 $\times$ 512  &50  &  19.86 & 3.27 \\ 
  & 192 $\times$ 512 & 100 & 18.87 & 6.20 \\ \midrule
   
\multirow{3}{*}{NuScenes-Occupancy} & 256 $\times$ 448 &20 & 15.86 & 1.76\\      
  & 256 $\times$ 448 &50 & 11.87 & 3.42 \\ 
  & 256 $\times$ 448 & 100 & 10.71 & 6.61 \\ \midrule
\end{tabular}
}
\caption{\bh{FID scores of the proposed method with different sampling steps. The evaluations are conducted on the NYUv2 test set, SemanticKITTI validation set and Nuscene-Occupancy validation set, respectively. }} 
\label{efficiency}
\vspace{-0pt}
\end{table}

\noindent\textbf{Efficiency Analyse.}
We report the running time and generation quality of several schemes across different datasets with our proposed OccScene on the NVIDIA A100 GPU, which are detailed in Table~\ref{efficiency}. It's worth noting that our method could effectively achieve compelling performance gains with acceptable time consumption. 
The results reveal that beyond 50 sampling steps, the marginal gains in effectiveness are minimal relative to the increased computational time. Consequently, we have selected 50 sampling steps as the default configuration, which provides an optimal balance between efficiency and effectiveness.\looseness=-1

\section{Conclusion}
In this paper, we propose {OccScene}, a unified framework that integrates fine-grained 3D perception and high-quality generation, resulting in mutual benefits with performance enhancements for both perception and generation. 
OccScene incorporates semantic occupancy within a joint-training diffusion framework and aligns occupancy with the diffusion latent using a Mamba-based Dual Alignment module.
Extensive experiments demonstrate that OccScene generates indoor and outdoor realistic 3D scenes. 
Furthermore, the framework significantly enhances the perception model, achieving state-of-the-art performance in the 3D semantic occupancy prediction task.\looseness=-1

\section*{Acknowledgments}
This work was supported in part by NSFC 62302246 and ZJNSFC under Grant LQ23F010008, and supported by the High Performance Computing Center at Eastern Institute of Technology, Ningbo, and Ningbo Institute of Digital Twin.

\bibliographystyle{IEEEtran}
\bibliography{egbib}

\begin{IEEEbiography}[{\includegraphics[width=1in,height=1.25in,clip,keepaspectratio]{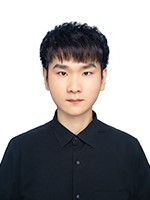}}]{Bohan Li} (Student Member, IEEE) received the B.E. degree from the School of Control Engineering, Northeastern University (NEU),
Shenyang, China, in 2019. He received the M.E. degree from the School of Control Science and Engineering, South China University of Technology
(SCUT), Guangzhou, China, in 2022.
He is currently pursuing the Ph.D. degree in Shanghai Jiao Tong University (SJTU) and Eastern Institute of Technology (EIT). His research interests include 3D visual perception, robotics, and multi-modality content generation.
\end{IEEEbiography}

\begin{IEEEbiography}[{\includegraphics[width=1in,height=1.25in,clip,keepaspectratio]{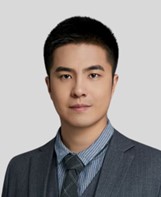}}]{Xin Jin} (Member, IEEE) has been a tenure track Assistant Professor with the Eastern Institute of Technology (EIT), Ningbo, China. He is also a Researcher at the Ningbo Institute of Digital Twin. He received his Ph.D. degree in Electronic Engineering and Information Science from the University of Science and Technology of China (USTC). His research interests include computer vision, intelligent media computing, and deep learning. He has over 10 granted patent applications, around 40 publications, and over 3,500 Google citations. He is an IEEE member, and reviewer of IEEE Transactions on Image Processing (TIP), IEEE Transactions on Multimedia (TMM), and IEEE Transactions on Circuits and Systems for Video Technology (TCSVT).
\end{IEEEbiography}

\begin{IEEEbiography}[{\includegraphics[width=1in,height=1.40in,clip,keepaspectratio]{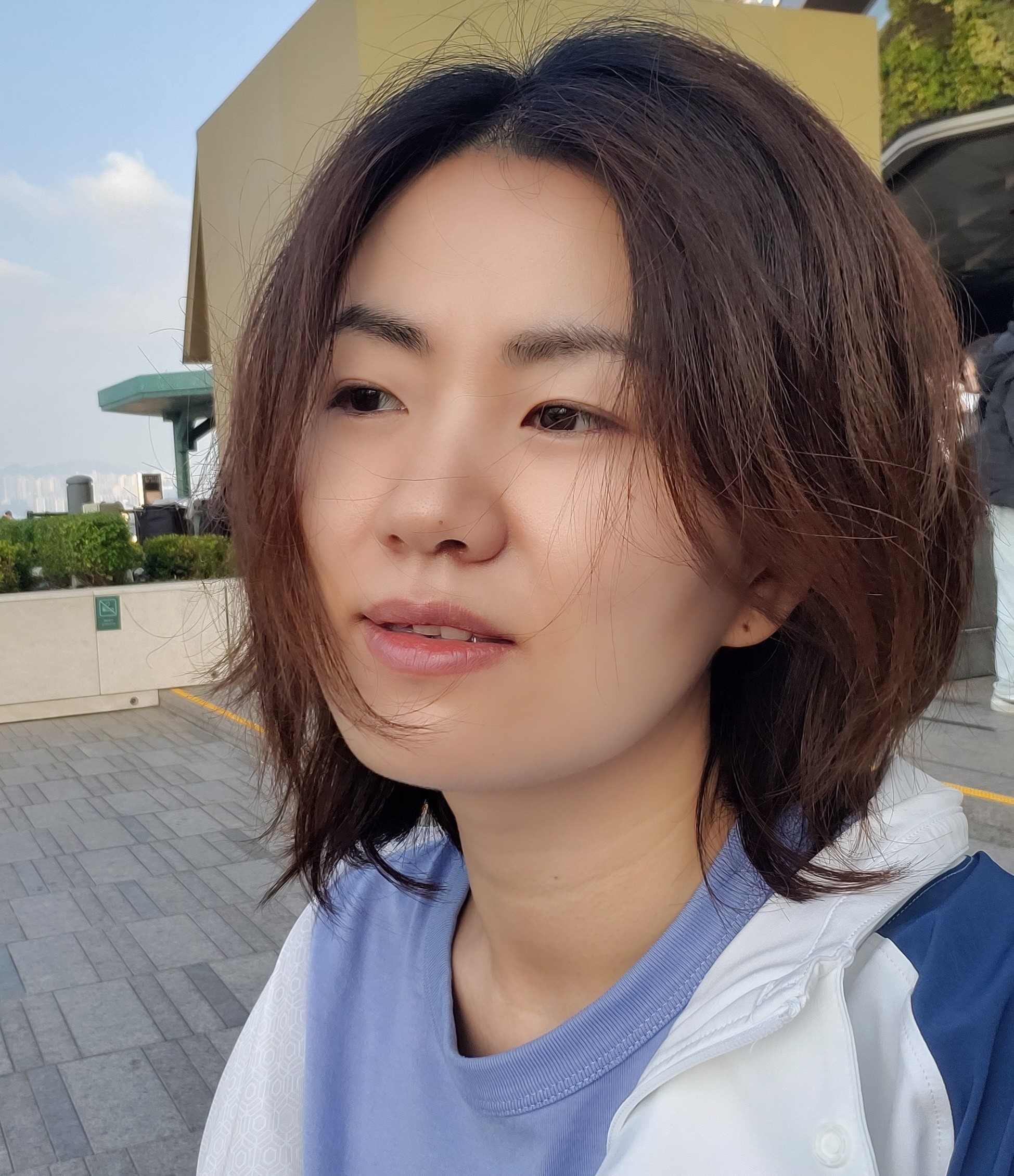}}]{Jianan Wang} is the Chief Researcher in AI cognition at Astribot. Previously, she received the bachelor’s degree from the Chinese University of Hong Kong, China. she received the MSc degree from the University of Oxford, UK. She has previously worked with DeepMind and the International Digital Economy Academy. 

Her research interests include computer vision, deep learning, and machine learning theory, with a recent focus on generative AI and robotics.
\end{IEEEbiography}

\begin{IEEEbiography}
[{\includegraphics[width=1.0in,height=1.1in,clip,keepaspectratio]{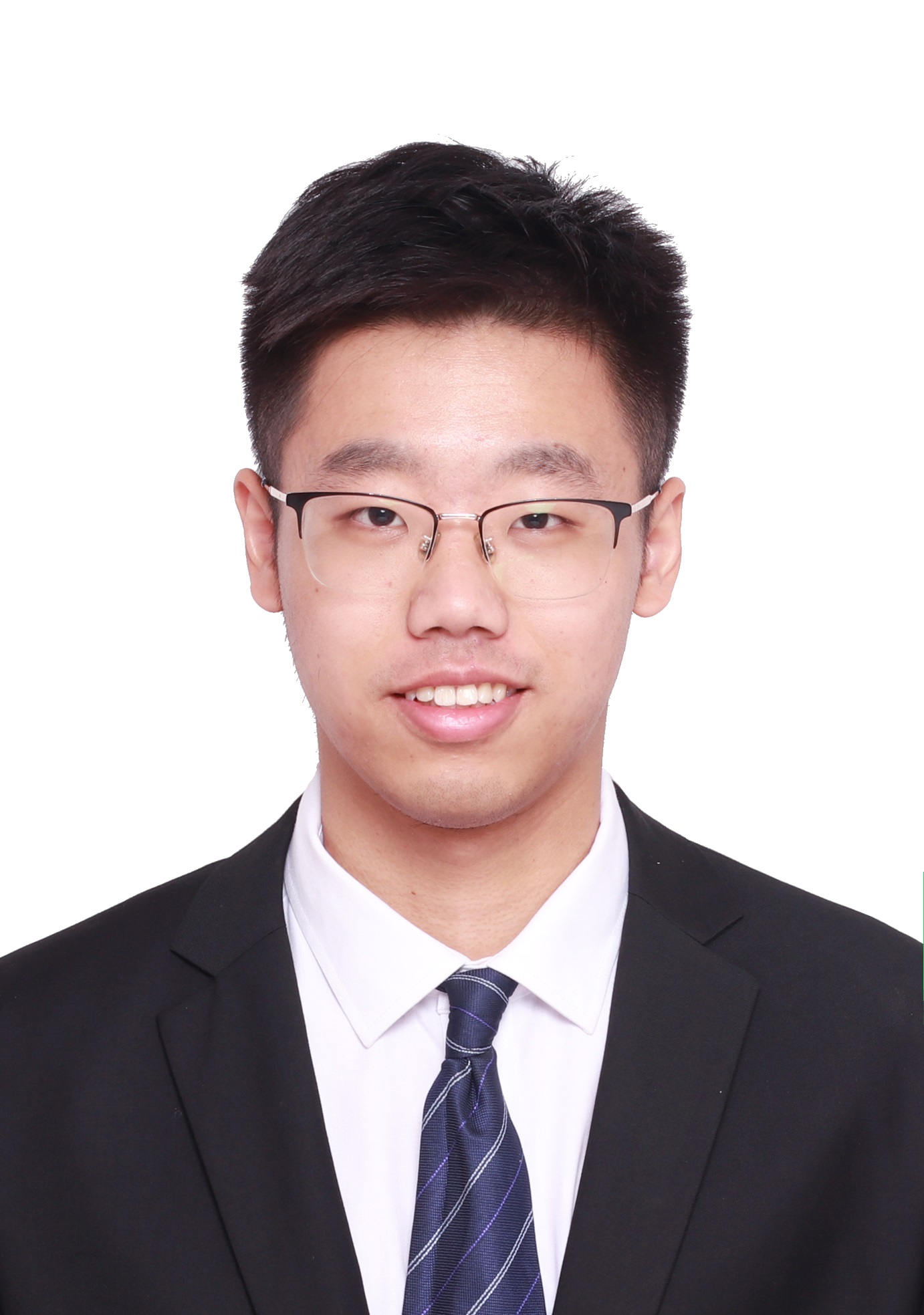}}]{Yukai Shi} (Student Member, IEEE) received the B.E. degree from the School of Artificial Intelligence, Xidian University (XDU), Xi’an, China, in 2022. 
He is currently pursuing the Ph.D. degree in Tsinghua University (THU), Beijing, China. He interned at the company of Astribot.
His research interests include 3D generation and video generation.
Furthermore, as a student member of IEEE, he serves as a reviewer for multiple computer Vinson conferences including CVPR, ACM MM, NeurIPS, ICLR.
\end{IEEEbiography}

\begin{IEEEbiography}[{\includegraphics[width=1in,height=1.25in,clip,keepaspectratio]{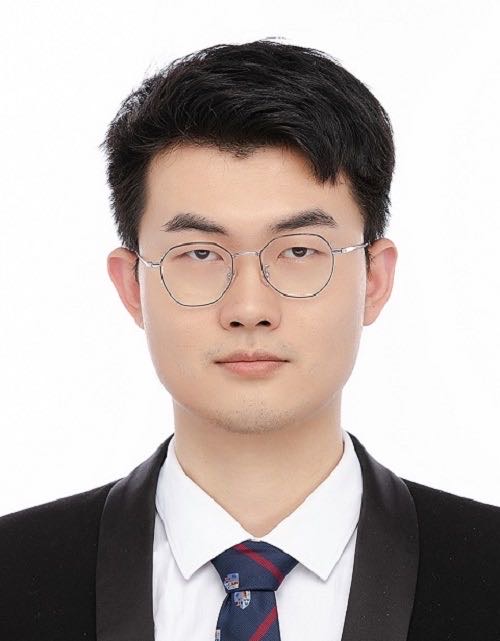}}]{Yasheng Sun} received the B.E. degree from Nanjing University of Aeronautics and Astronautics, Nanjing, China, in 2017. He received the M.E. degree from the School of Mechanical Engineering, Shanghai Jiao Tong University, Shanghai, China, in 2020. 
He received the Ph.D. degree in Computer Science from the School of Computing, Tokyo Institute of Technology, Japan, in 2024. He interned at the company of Astribot.
His current research interest includes cross-modal generation, stable diffusion model and its application in computer vision.
\end{IEEEbiography}

\begin{IEEEbiography}[{\includegraphics[width=1in,height=1.25in,clip,keepaspectratio]{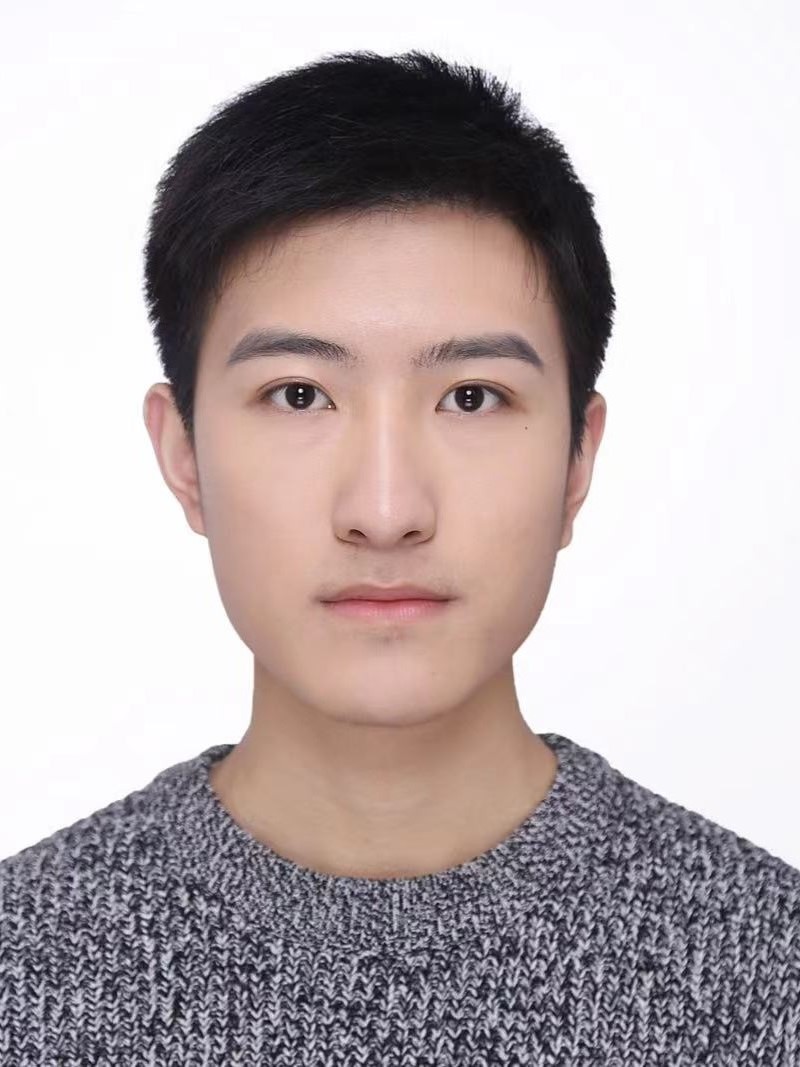}}]{Xiaofeng Wang} received the B.E. degree from the School of Automation, Nanjing University of Science and Technology (NJUST), Nanjing, China, in 2020. He is currently pursuing the Ph.D. degree in Institute of Automation, Chinese Academy of Science (CASIA), Beijing, China. He interned at the company of Astribot. 
His current research areas include 3D perception and video generation. He has co-authored 10+ journal and conference papers mainly on computer vision autonomous-driving problems, including CVPR, ECCV, ICCV, AAAI, and ICLR.
\end{IEEEbiography}

\begin{IEEEbiography}
[{\includegraphics[width=1.0in,height=1.1in,clip,keepaspectratio]{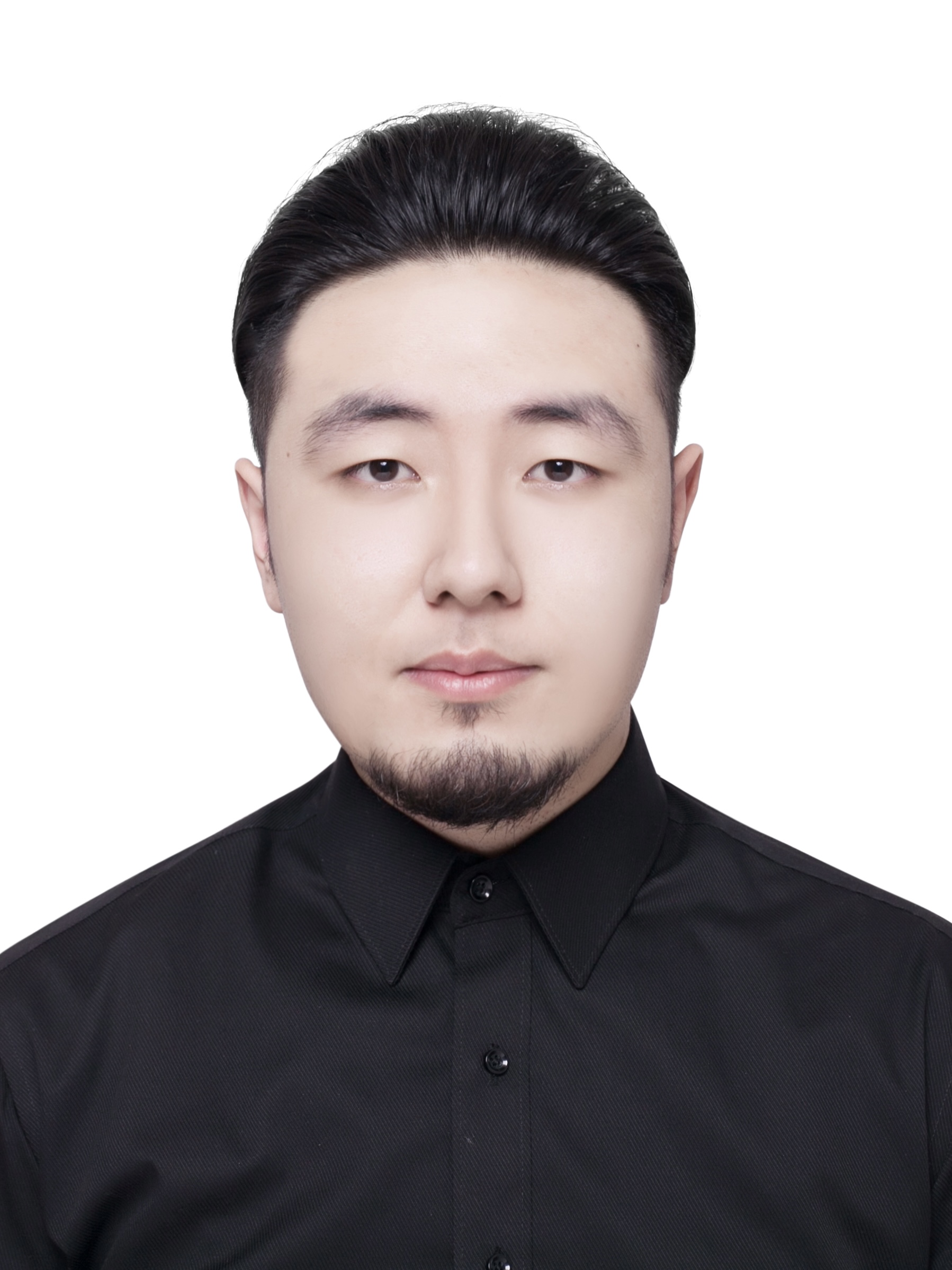}}]{Zhuang Ma} received the B.E. degree from the University of Plymouth, UK, in 2020. He received the MSc degree from the University of Birmingham, UK, in 2021. He is currently an engineer at PhiGent Robotics, Beijing, China. 

His current research interests include 2D and 3D visual perception, robotics, and multi-modality content generation. He serves as a reviewer for multiple computer Vinson conferences including CVPR, ICCV, AAAI.
\end{IEEEbiography}

\begin{IEEEbiography}
[{\includegraphics[width=1.0in,height=1.1in,clip,keepaspectratio]{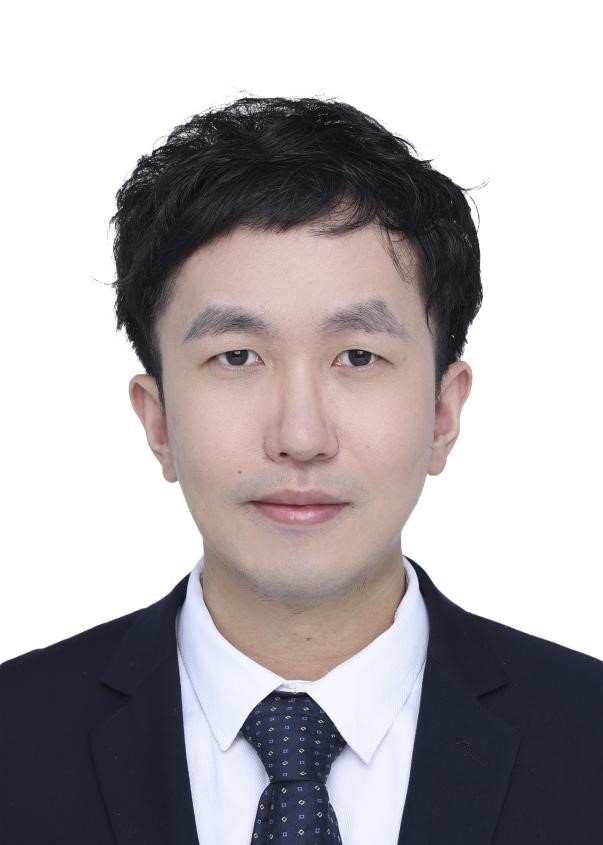}}]{Baao Xie} (Member, IEEE) obtained his B.E. degree from Northeastern University, China, and both his M.S. and Ph.D. degrees from Loughborough University, UK, in 2021. Subsequently, he engaged in postdoctoral research at the Eastern Institute of Technology (EIT) and Tian Jing University under the supervision of Wenjun Zeng (the academician of the Canadian Academy of Engineering, the Vice President for EIT, the founding Executive Director of Ningbo Institute of Digital Twin (IDT), IEEE Fellow). His research interests include 3D reconstruction, disentangled representation learning (DRL), Neural Radiance Fields, Gaussian Splatting, Graph, Multimodal Large Models. He has disseminated his scholarly work on 3D reconstruction and (DRL) through publications in top venues including ICCV, CVPR, NeurIPS, ECCV, CMPB, and etc, with several relevant patent applications. Furthermore, as a member of IEEE, he serves as a reviewer for multiple computer Vinson conferences/journals, contributing to the advancement of the academic community in his fields of expertise.
\end{IEEEbiography}

\begin{IEEEbiography}[{\includegraphics[width=1in,height=1.25in,clip,keepaspectratio]{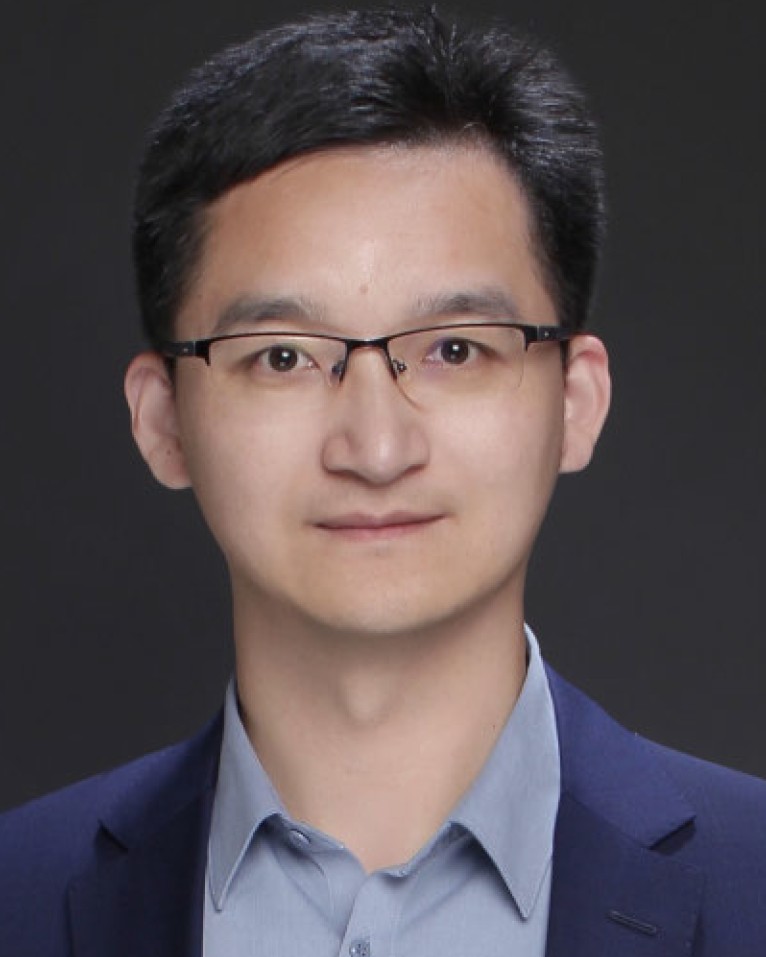}}]{Chao Ma} (Member, IEEE) received the Ph.D. degree from Shanghai Jiao Tong University, Shanghai, China, in 2016. He was sponsored by the China Scholarship Council as a Visiting Ph.D. Student at the University of California at Merced, Merced, CA, USA, from Fall 2013 to Fall 2015. He was a Research Associate with the School of Computer Science, The University of Adelaide, Adelaide, SA, Australia, from 2016 to 2018. He is currently an Associate Professor at Shanghai Jiao Tong University. His research interests include computer vision and machine learning.
\end{IEEEbiography}

\begin{IEEEbiography}[{\includegraphics[width=1in,height=1.25in,clip,keepaspectratio]{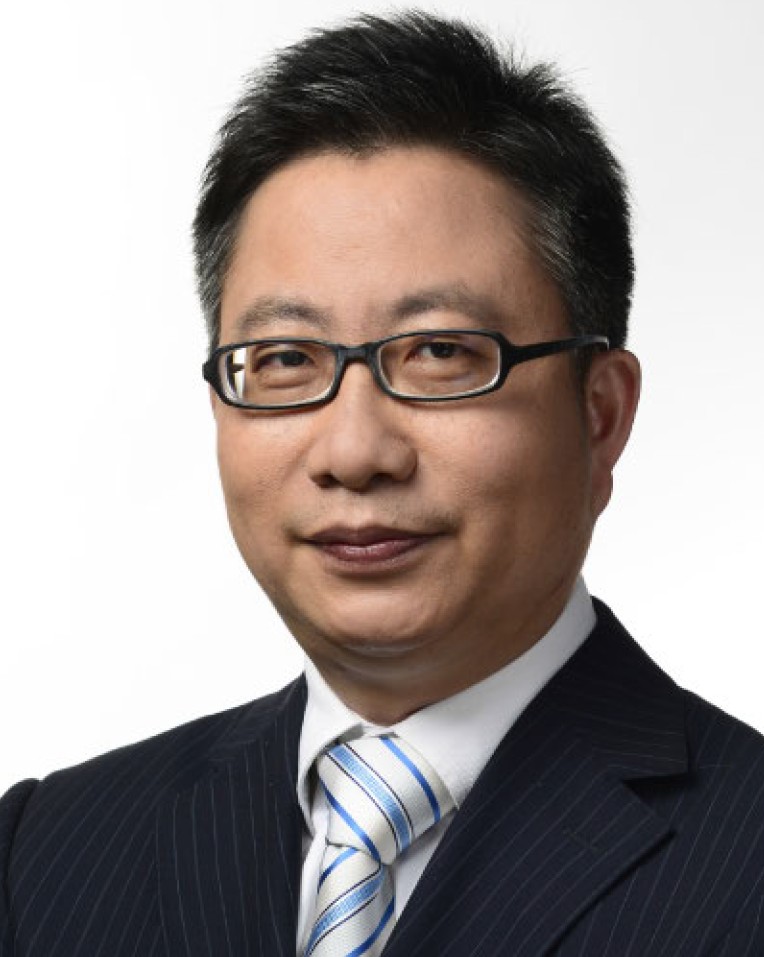}}]{Xiaokang Yang} (Fellow, IEEE) received the B.S. degree from Xiamen University, Xiamen, China, in 1994, the M.S. degree from the Chinese Academy of Sciences, Shanghai, China, in 1997, and the Ph.D. degree from Shanghai Jiao Tong University, Shanghai, in 2000. From September 2000 to March 2002, he worked as a Research Fellow with the Centre for Signal Processing, Nanyang Technological University, Singapore. From April 2002 to October 2004, he was a Research Scientist at the Institute for Infocomm Research (I2R), Singapore. From August 2007 to July 2008, he visited the Institute for Computer Science, University of Freiburg, Breisgau, Germany, as an Alexander von Humboldt Research Fellow. He is currently a Distinguished Professor at the School of Electronic Information and Electrical Engineering, Shanghai Jiao Tong University. He has published over 200 refereed articles and has filed 60 patents. His research interests include image processing and communication, computer vision, and machine learning. Dr. Yang received the 2018 Best Paper Award of IEEE TRANSACTIONS ON MULTIMEDIA. He is an Associate Editor of IEEE TRANSACTIONS ON MULTIMEDIA and a Senior Associate Editor of IEEE SIGNAL PROCESSING LETTERS.
 \end{IEEEbiography}

\begin{IEEEbiography}[{\includegraphics[width=1in,height=1.25in,clip,keepaspectratio]{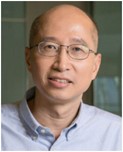}}]{Wenjun Zeng} (Fellow, IEEE) received the B.E. degree from Tsinghua University, Beijing, China,
in 1990, the M.S. degree from the University of Notre Dame, Notre Dame, IN, USA, in 1993, and the Ph.D. degree from Princeton University, Princeton, NJ, USA, in 1997. He has been a Chair Professor and the Vice President for Research at the Eastern Institute for Advanced Study (EIAS) / Eastern Institute of Technology (EIT), Ningbo, China, since October 2021. He is also the founding Executive Director of the Ningbo Institute of Digital Twin. He was a Sr. Principal Research Manager and a member of the Senior Leadership Team at Microsoft Research Asia, Beijing, from 2014 to 2021, where he led the video analytics research empowering the Microsoft Cognitive Services, Azure Media Analytics Services, Office, and Windows Machine Learning. He was with University of Missouri, Columbia, MO, USA from 2003 to 2016, most recently as a Full Professor. Prior to that, he had worked for PacketVideo Corp., Sharp Labs of America, Bell Labs, and Panasonic Technology. He has contributed significantly to the development of international standards (ISO MPEG, JPEG2000, and OMA).
Dr. Zeng is on the Editorial Board of the International Journal of Computer Vision. He was an Associate Editor-in-Chief of the IEEE Multimedia Magazine and an Associate Editor of the IEEE TRANSACTIONS ON CIRCUITS AND SYSTEMS FOR VIDEO TECHNOLOGY, IEEE TRANSACTIONS ON INFORMATION FORENSICS AND SECURITY, and IEEE TRANSACTIONS
ON MULTIMEDIA (TMM). He was on the Steering Committee of IEEE TRANSACTIONS ON MOBILE COMPUTING and IEEE TMM. He served as the Steering Committee Chair of IEEE ICME in 2010 and 2011, and has served as the General Chair or TPC Chair for several IEEE conferences (e.g., ICME’2018, ICIP’2017). He was the recipient of several best paper awards.
\end{IEEEbiography}


\end{document}